\documentclass{article}

\PassOptionsToPackage{numbers, compress}{natbib}

 \usepackage[dblblindworkshop, final]{neurips_2026}
\workshoptitle{Geometric Distributional Deep Learning (Oral)}

\usepackage{hyperref}
\usepackage{booktabs}
\usepackage{microtype}

\usepackage{amssymb}
\usepackage{mathtools}
\usepackage{bm}
\usepackage{pifont}
\usepackage{makecell}
\usepackage[table]{xcolor}
\usepackage{nicematrix}
\usepackage{multirow}
\usepackage[ruled,vlined,linesnumbered]{algorithm2e}
\DontPrintSemicolon
\usepackage{caption}
\usepackage{wrapfig}
\usepackage{enumitem}
\setlist[itemize]{topsep=3pt, itemsep=3pt, parsep=0pt, partopsep=0pt}
\setlist[enumerate]{topsep=3pt, itemsep=3pt, parsep=0pt, partopsep=0pt}
\usepackage{tocloft} 
\usepackage{etoc}

\definecolor{teal_main}{HTML}{48C9B0}
\definecolor{blue_dark}{HTML}{34495E}
\definecolor{linkblue}{HTML}{195dcf}
\hypersetup{
  colorlinks=true,
  urlcolor=linkblue,
  linkcolor=black,
  citecolor=black
}

\newcommand{\R}{\mathbb{R}}

\newcommand{\x}{\mathbf{x}}

\newcommand{\e}{\bm{\epsilon}}
\newcommand{\z}{\mathbf{z}}
\newcommand{\G}{\mathbf{G}}

\title{Efficient Graph Generation\\via Direct Prediction and Flow Matching}

\author{%
  Susie Lu\\
  MIT
}

\begin{document}
\raggedbottom

\maketitle

\vspace{-6pt}

\begin{abstract}
\vspace{-6pt}
Generative modeling of graph-structured data is crucial for tasks ranging from drug discovery to social network simulation. Among these models, denoising diffusion models have achieved great success in graph generation by learning to progressively reverse a process that adds noise to the original graph. However, the standard noise-prediction approach of diffusion models is suboptimal for graph data. The goal for a graph generative model is to learn the clean graphs’ topological properties, such as connectivity and degree distribution. Because a diffusion model that predicts noise does not explicitly learn these topological properties, it is challenging for the model to output graphs with the desired structural statistics. To address this challenge, we introduce \textbf{Direct Graph Flow Matching (DiGFM)}, a novel graph transformer model guided by two goals: predict clean graphs and improve sampling efficiency. Distinct from the prevailing diffusion approach, DiGFM employs a continuous flow-matching paradigm and integrates direct graph prediction. Specifically, DiGFM maps the prior noise distribution to the clean graph distribution via a multi-step process: the model repeatedly predicts the underlying clean graph, and a transformation is employed to convert the model output to the velocity vector that points in the direction toward the clean graph distribution. This design enables DiGFM to generate high-quality samples using only 2.5\% to 15.6\% of the steps required by diffusion-based models, which leads to a 5.3$\times$ to 257$\times$ speedup in wall-clock inference time. Experiments demonstrate that DiGFM outperforms or matches prior state-of-the-art models across general graph benchmarks and molecular datasets, generating graphs with strong adherence to ground-truth structural statistics at significantly faster inference speeds.
\end{abstract}

\etocdepthtag.toc{main}
\vspace{-5pt}
\section{Introduction}
\vspace{-3pt}

Graph generative models have advanced rapidly, with diffusion models achieving great success across diverse domains such as molecular generation \cite{irwin2024efficient}, protein design \cite{watson2023novo}, neural architecture search \cite{asthana2024multi}, social network analysis \cite{liu2023survey}, and combinatorial optimization \cite{sun2023difusco}. These models generate graphs by learning to reverse a forward process that progressively corrupts the original data with noise \cite{ho2020denoising,song2021score}. Typically, this reversal is accomplished by having the network predict either the added noise or the original clean graph.

The prevailing approach is noise prediction \cite{ho2020denoising}, but it is suboptimal for graph data for two reasons. First, the primary goal of a graph generative model is to learn graph topological properties, such as connectivity and degree distribution. Diffusion models that predict noise do not explicitly reconstruct these properties, making it challenging for them to output graphs with faithful structural statistics. Instead, a diffusion model that directly predicts the original clean graph can generate more accurately. Second, noise prediction suffers from a curse of dimensionality rooted in the manifold hypothesis \cite{chapelle2006semi,li2025back}. This hypothesis states that real-world data lies roughly on a low-dimensional manifold. Since clean graphs possess specific topological structures, they approximately lie on a low-dimensional manifold. In contrast, noise (e.g., Gaussian noise in diffusion) is off-manifold and distributed across the entire high-dimensional space. Because our model directly predicts clean graphs (low-dimensional) rather than noise (high-dimensional), it learns the topological properties of the graph more easily and achieves significantly greater sampling efficiency.

While the approach of predicting clean graphs has been explored by GruM \cite{jo2024graph}, which predicts a weighted mean of clean graph structures, its underlying diffusion mixture approach presents limitations. GruM relies on Ornstein-Uhlenbeck processes \cite{doob1984classical}, which induce curved, stochastic trajectories from noise to data. As a result, it requires a large number of sampling steps (e.g., 1,000) during inference to accurately approximate these curved trajectories and maintain generation quality.

To overcome these limitations, we introduce \textbf{Direct Graph Flow Matching (DiGFM)}, a novel generative model that directly predicts clean graphs and leverages the recent flow matching paradigm \cite{lipman2023flow} to significantly improve sampling efficiency. Unlike prior work, DiGFM decouples the model's output space (clean graph) from the loss space (velocity in flow matching). Specifically, given a noisy graph, DiGFM directly outputs a clean graph prediction; this prediction is transformed into a velocity vector, which an Ordinary Differential Equation (ODE) solver uses to take a step toward the next, less noisy version of the graph.

Our experiments demonstrate that DiGFM outperforms or matches existing generative models across diverse datasets while requiring only a fraction of their sampling steps. On general graph benchmarks (Community \cite{you2018graphrnn}, Ego \cite{sen2008collective}, Stochastic Block Model \cite{martinkus2022spectre}, and Enzymes \cite{schomburg2004brenda}), DiGFM outperforms prior models on the majority of the metrics, which cover both classical structural metrics (degree, clustering, orbit MMDs) and neural-based metrics (FID and RBF MMD). Notably, DiGFM achieves lower values for the average maximum mean discrepancy and neural-based metrics using only 15 to 20 sampling steps---just 2.5\% to 15.6\% of the steps required by previous state-of-the-art models. On molecular generation, DiGFM achieves a validity of 99.82\% on QM9 \cite{ramakrishnan2014quantum} and 98.73\% on ZINC250k \cite{irwin2012zinc}, which indicates that almost all of the generated molecules are chemically valid without post hoc correction. Furthermore, DiGFM naturally supports conditional generation, and it achieves state-of-the-art performance on generating QM9 molecules conditioned on the dipole moment and on the highest occupied molecular orbital (HOMO).

Our main contributions can be summarized as follows:
\vspace{-5pt}
\begin{itemize}
    \item We introduce DiGFM, a generative model that directly predicts clean graphs in order to capture topological properties better than standard noise-prediction models.
    \item We develop a framework that integrates direct graph prediction with the flow matching paradigm.
    \item Our model outperforms or matches prior state-of-the-art models on general graph benchmarks and molecular datasets while achieving a 5.3$\times$ to 257$\times$ wall-clock inference speedup by using fewer sampling steps.
\end{itemize}
\vspace{-5pt}
Our code is available at \href{https://github.com/Susie-Lu/DiGFM}{this link}.

\vspace{-6pt}
\section{Related Work}
\vspace{-6pt}

Recent advancements in graph generative modeling are predominantly driven by diffusion and flow-based frameworks. These models can be categorized into continuous and discrete approaches. Continuous models operate on node and edge attribute matrices in continuous space (i.e., $\R^{n \times d_x}$ and $\R^{n \times n \times d_e}$), aiming to denoise initial Gaussian matrices into the target graph's attributes. In contrast, discrete models are designed specifically for graphs with categorical node and edge attributes. They aim to denoise the prior probability distributions to recover the actual probability distributions of discrete node and edge types.

\textbf{Continuous Graph Diffusion/Flow Models.} Table \ref{tab:con-pred} summarizes the prediction spaces of the continuous models. Graph models based on the standard Denoising Diffusion Probabilistic Models (DDPM) framework \cite{ho2020denoising} predict the added noise at each timestep. Similarly, score-based models \cite{song2021score} predict the score, which is proportional to the negative of the noise. Models based on diffusion bridges \cite{peluchetti2023non} and flow matching \cite{lipman2023flow,liu2023flow,albergo2023building} predict the drift or velocity, which mathematically represents the difference between the clean graph and noise.

Continuous graph models operate in either the original graph space or in a compressed latent space via an autoencoder. While latent space models \cite{zhou2024unifying, yang2024graphusion, siraudin2026principled} offer a unique paradigm, their reliance on autoencoders can introduce reconstruction artifacts. Specifically, the autoencoder may be unable to distinguish between certain types of vertices \cite{siraudin2026principled}, and strict structural properties (e.g., planarity) can be violated through a single edge reconstruction error. Since our primary objective is to generate topological structures with high fidelity, DiGFM operates directly in the graph space to bypass potential autoencoder bottlenecks.

Among continuous graph models, only GruM \cite{jo2024graph} and our DiGFM directly predict the clean graph rather than noise, velocity, or related quantities. This direct prediction encourages both models to capture target structural statistics more accurately. However, GruM relies on a diffusion mixture framework \cite{peluchetti2023non}, which induces highly curved sampling trajectories. By leveraging the flow matching framework, inspired by recent success in direct pixel-space image generation \cite{li2025back}, DiGFM is able to generate high-quality graphs with significantly fewer sampling steps than GruM.

\begin{table}[h!]
\vspace{-5pt}
\centering
\caption{Prediction spaces of continuous graph diffusion/flow models.}
\vspace{2pt}
\label{tab:con-pred}
\tiny
\begin{tabular}{cccc}
\toprule
Framework & Graph Models & Pred. Space & \makecell{Predicts \\ Clean Graph?} \\
\hline
\rowcolor[gray]{0.9} \multicolumn{4}{l}{\textit{Graph Space Models}}  \\
DDPM & ConGress \cite{vignac2023digress} & Noise & \textcolor{red}{\ding{55}} \\
Score-based & EDP-GNN \cite{niu2020permutation}, GDSS \cite{jo2022score} & Score & \textcolor{red}{\ding{55}} \\
Diffusion Bridge & Bridge \cite{wu2022diffusion} & Drift & \textcolor{red}{\ding{55}} \\
Diffusion Mixture & GruM \cite{jo2024graph} & Graph & \textcolor{green}{\ding{51}} \\
\textbf{Flow-based} & \textbf{DiGFM (ours)} & \textbf{Graph} & \textcolor{green}{\ding{51}} \\
\hline
\rowcolor[gray]{0.9} \multicolumn{4}{l}{\textit{Latent Space Models}}  \\
DDPM & LGD \cite{zhou2024unifying} & Noise & \textcolor{red}{\ding{55}} \\
Score-based & Graphusion \cite{yang2024graphusion}, LGD \cite{zhou2024unifying} & Score & \textcolor{red}{\ding{55}} \\
Flow-based & LG-Flow \cite{siraudin2026principled} & Velocity & \textcolor{red}{\ding{55}} \\
\bottomrule
\end{tabular}
\vspace{2pt}
\end{table}

\textbf{Discrete Graph Diffusion/Flow Models.} Table \ref{tab:di-pred} summarizes the prediction spaces of discrete models, which are an important parallel line of research to continuous models. Notably, these models predict the clean graph, since
they output categorical probability distributions for every node and edge type and are trained to match the clean graph's node and edge categories via cross-entropy loss.

While the discrete approach is prevalent among graph diffusion/flow models, we choose a continuous framework for DiGFM due to three primary advantages. First, the continuous approach naturally allows our model to generate graphs with both continuous and discrete attributes; in contrast, discrete approaches require added architectural components, e.g. \cite{vignac2023midi}, to generate both continuous and categorical features. Second, operating in continuous space allows DiGFM to straightforwardly adopt advancements in image-domain diffusion/flow models, such as integrating classifier-free guidance \cite{ho2021classifier} for conditional molecular generation (e.g., generating QM9 molecules with specific dipole moments). Third, in continuous space, we can leverage numerical ODE solvers \cite{karras2022elucidating} with large step sizes to approximate the trajectory from noise to the clean graph. In practice, this leads to greater sampling efficiency compared to the standard Markov chains \cite{campbell2024generative} used in discrete flow models.

\begin{table}[h!]
\vspace{-5pt}
\centering
\caption{Prediction spaces of discrete graph diffusion/flow models.}
\vspace{2pt}
\label{tab:di-pred}
\tiny
\begin{tabular}{lccc}
\toprule
Framework & Graph Models & Pred. Space & \makecell{Predicts \\ Clean Graph?} \\
\hline
\rowcolor[gray]{0.9} \multicolumn{4}{l}{\textit{Graph Space Models}}  \\
Discrete DDPM {\tiny (discrete $t$)} & DiGress \cite{vignac2023digress} & Graph & \textcolor{green}{\ding{51}} \\
Discrete DDPM {\tiny (continuous $t$)} & DisCo \cite{xu2024discrete} & Graph & \textcolor{green}{\ding{51}} \\
Discrete Flow-based & DeFoG \cite{qin2025defog} & Graph & \textcolor{green}{\ding{51}} \\
\bottomrule
\end{tabular}
\vspace{-5pt}
\end{table}
\section{Direct Graph Flow Matching}\label{sec:DiGFM}
\vspace{-5pt}

In this section, we present our Direct Graph Flow Matching (DiGFM) model. We first formulate the graph generation problem, then introduce our novel integration of direct graph prediction into the flow matching framework, and finally explain the graph-specific architectural designs required to make this approach work well.

\vspace{-4pt}
\subsection{Problem Formulation}\label{sec:method-problem}
\vspace{-4pt}

The objective of graph generation is to learn a generative model that maps a prior noise distribution $p_{\mathrm{noise}}$ to a target data distribution $p_{\mathrm{data}}$ of realistic graphs. Once trained on a given dataset, this model can be efficiently sampled to generate diverse, novel graphs that share the structural and topological properties of the original graphs.

Consider an $n$-vertex graph $\G$, which can be undirected or directed. We represent $\G$ by its node attribute matrix $\bm{X} \in \R^{n \times d_x}$ and its edge attribute matrix $\bm{E} \in \R^{n \times n \times d_e}$, where $d_x$ and $d_e$ are the dimensionalities of the node and edge attributes, respectively. These attributes can be either continuous or discrete. Continuous attributes, such as edge weights, are represented by their direct values within the attribute matrices. Discrete attributes are encoded using a one-hot representation within these matrices. Thus, the learning objective is to approximate a continuous vector field that transports initial Gaussian noise matrices $(\bm{X}_0, \bm{E}_0)$ into realistic clean attribute matrices $(\bm{X}_1, \bm{E}_1)$.

\vspace{-4pt}
\subsection{Direct Graph Generation via Flow Matching}\label{sec:method-flow}
\vspace{-4pt}

To address the problem of graph generation, we integrate direct, clean graph prediction into the flow matching framework.

\textbf{Flow Matching Framework.} Let $\x = (\bm{X}, \bm{E})$ denote the clean graph sampled from $p_{\mathrm{data}}$, where $\bm{X}$ and $\bm{E}$ are the node and edge attribute matrices, respectively. Similarly, let $\e = (\e_X, \e_E)$ denote Gaussian noise sampled from $p_{\mathrm{noise}}$ with the same tensor shapes. We define a forward process that constructs a noisy sample $\z_t = (\bm{X}_t, \bm{E}_t)$ at any time step $t \in [0, 1]$ via a linear interpolation:
$$\z_t = t \x + (1 - t) \e.$$
Here, all scalar operations apply component-wise, meaning $\bm{X}_t = t \bm{X} + (1 - t) \bm{\epsilon}_X$ and $\bm{E}_t = t \bm{E} + (1 - t) \bm{\epsilon}_E$. Under this formulation, $\z_t$ represents pure noise at time $t=0$ and the clean graph at time $t=1$. The flow velocity $\mathbf{v} = (\mathbf{v}_X, \mathbf{v}_E)$ is the time-derivative of $\z_t$: specifically, $\mathbf{v} = \frac{d}{dt} \z_t = \x - \e$. Flow-based methods \cite{lipman2023flow,liu2023flow,albergo2023building} train a neural network $\mathbf{v}_\theta(\z_t, t)$ to predict this velocity. Since $\z_t$ is a tuple, the network takes both $\bm{X}_t$ and $\bm{E}_t$ as input and outputs predicted velocities for both nodes and edges. The model is trained by minimizing the loss function:
\begin{equation}\label{eq:loss}
\mathcal{L} = \mathbb{E}_{t, \x, \e} \; \| \mathbf{v}_\theta(\z_t, t) - (\x - \e) \|^2.
\end{equation}
where the squared difference decomposes into the sum of the squared differences over the node and edge components. Once the model has been trained, samples can be generated by solving the ODE $\frac{d}{dt} \z_t = \mathbf{v}_{\theta}(\z_t, t)$, starting from noise at $t=0$ and ending at a generated graph at $t=1$. This can be done by numerical methods such as the Euler or Heun ODE solvers.

\textbf{Adapting to Direct Clean Graph Prediction.} While predicting velocity or noise is standard practice for flow matching and diffusion models (which were initially developed for image generation), graph topologies are very sensitive to noise and intermediate continuous states. As noted earlier, by predicting the clean graph instead, the model is forced to explicitly reconstruct topological properties. Further, it avoids the inefficiency of predicting high-dimensional, off-manifold targets like noise $\e$ or velocity $\mathbf{v}$, since velocity $\mathbf{v}=\x-\e$ inherits the high dimensionality of noise. Thus, we train DiGFM to directly generate the clean graph $\x$.

To do so, we decouple the loss space ($\mathbf{v}$-space) from the model's output space ($\x$-space). Given the constraints $\z_t = t \x + (1 - t) \e$ and $\mathbf{v} = \x - \e$, we can solve for the velocity in terms of the clean graph $\x$ and the current noisy state $\z_t$, yielding $\mathbf{v} = \frac{1}{1 - t}(\x - \z_t)$. This allows us to define the transformation:
\begin{equation}\label{eq:x-to-v}
\boxed{\mathbf{v}_\theta(\z_t, t) \coloneqq \tfrac{1}{1 - t} (\x_\theta(\z_t, t) - \z_t).}
\end{equation}
Thus, DiGFM can directly predict the clean graph $\x_\theta$; this predicted clean graph is transformed into velocity $\mathbf{v}_\theta$ via equation \eqref{eq:x-to-v}; finally, the model is optimized using the flow matching objective \eqref{eq:loss}.

Substituting equation \eqref{eq:x-to-v} into \eqref{eq:loss} reveals an alternative perspective on DiGFM's loss function:
$$\mathcal{L} = \mathbb{E}_{t, \x, \e} \; \tfrac{1}{(1-t)^2} \Vert{} \x_\theta(\z_t, t) - \x \Vert{}^2.$$
This expression can be interpreted as a reweighted form of L2 loss measuring the deviation between the model's predicted clean graph and the ground-truth clean graph. The factor $\frac{1}{(1-t)^2}$ assigns smaller weights near $t=0$ (where inferring the clean graph from pure noise is extremely difficult) and larger weights as $t \to 1$ (where the network must perfect the final graph topology).

\vspace{-3pt}
\subsection{Graph-Tailored Network Architecture and Inductive Biases}\label{sec:method-arch}
\vspace{-3pt}
DiGFM uses a graph transformer \cite{vignac2023digress,dwivedi2021generalization} as its backbone model. Specifically, the model takes the noisy graph $\G_t = (\bm{X}_t, \bm{E}_t)$ and time step $t$ as input, and it outputs the predicted clean graph $\G_1$.

To adapt this architecture for direct flow-based graph prediction, we introduce several crucial graph-specific inductive biases (details in Appendix \ref{app:details}):
\vspace{-5pt}
\begin{itemize}
\item \textbf{Output Bounding:} Since the model's output $\x_\theta(\z_t, t)$ should be a clean graph, we apply a sigmoid activation to the predicted edge matrix directly after the final transformer block. This bounds the model predictions to the interval $[0, 1]$, which helps prevent out-of-distribution values before converting to velocity via equation \eqref{eq:x-to-v}.
\item \textbf{Higher-Order Structural Features:} To help the model reconstruct the overall graph topology, we concatenate higher-order powers of the edge matrix (e.g., $\bm{E}_t$ and $\bm{E}_t^2$) during the forward pass, which provides information about multi-hop vertex neighborhoods.
\item \textbf{Laplacian Spectral Node Features:} For graph datasets without intrinsic node attributes, we define the clean target node attribute matrix $\bm{X}\in\mathbb{R}^{n\times k}$ as the top $k$ Laplacian eigenvectors computed from the ground-truth clean graph. DiGFM jointly predicts these spectral node features $\bm{X}$ and the clean edge matrix $\bm{E}$, which provides additional structural supervision when generating the target graph topology.
\end{itemize}

\vspace{-5pt}

\subsection{Training and Sampling Pipeline}\label{sec:method-train-sample}

\textbf{Training Algorithm.} The pseudocode of the training algorithm is in Algorithm \ref{alg:train}. For undirected graph generation, we sample the noise for the adjacency matrix to be symmetric with diagonal entries equal to zero. This ensures that the initial noise graph $\G_0$ is undirected and has no self-loops. Also, to prevent division by zero in line 7 of the algorithm, we cap the flow matching time step $t$ to be at most $(K_{\text{max}}-1)/K_{\text{max}}$, where $K_{\text{max}}$ denotes the maximum intended number of sampling steps at inference. With our default $K_{\text{max}}=100$, $t$ is clipped at 0.99. This numerical bounding does not affect the final generated graphs, because during inference, our $K$-step ODE solver only evaluates the velocity field at time steps $t \in \{0, 1/K, \dots, (K-1)/K\}$. Thus, it never queries $t > (K_{\text{max}}-1)/K_{\text{max}}$, so the sampling trajectory and output graph are unaffected (see Appendix \ref{app:ablate} for empirical validation).

To stabilize training, we apply an adaptive Exponential Moving Average (EMA) \cite{karras2022elucidating} to the model weights. We scale the EMA decay rate based on the number of processed graphs; this allows the model weights to update rapidly early in training and change more slowly as training progresses. More details about the EMA and Algorithm \ref{alg:train} are in Appendix \ref{app:details}.

\textbf{Sampling Algorithm.} To generate new graphs using DiGFM, we sample initial noise $(\bm{X}_0, \bm{E}_0)$ and solve the ODE $\frac{d}{dt} (\bm{X}_t, \bm{E}_t) = \mathbf{v}_\theta\left((\bm{X}_t, \bm{E}_t), t \right)$ from $t=0$ to $t=1$. At each integration step, we predict $\x_\theta$, convert it to $\mathbf{v}_\theta$ via equation \eqref{eq:x-to-v}, and step forward using an Euler solver (Algorithm \ref{alg:sample}). Since DiGFM operates in continuous time, the number of sampling steps $K$ can be flexibly adjusted. Finally, continuous representations of discrete categories are mapped back to categorical labels via thresholding (e.g., quantize the values to $1$ if $>0.5$, else $0$, for a binary adjacency matrix).
\vspace{-6pt}

\begin{minipage}[t]{0.48\textwidth}
\begin{algorithm}[H]
\caption{DiGFM Training}
\label{alg:train}

\textbf{Input:} Model $x_\theta(\G_t, t)$

\For{\emph{\textbf{each step}}}{
    Sample graph $\G_1$ from the training set \;
    $n \leftarrow$ number of nodes of $\G_1$ \;
    Sample $t \sim \mathcal{U}[0, 1]$ and $\G_0 \sim \mathcal{N}(0, I_n)$\;
    \BlankLine
    $\G_t \leftarrow t \cdot \G_1 + (1-t) \cdot \G_0$\;
    $\mathbf{v} \leftarrow (\G_1 - \G_t) / (1 - t)$\;
    \BlankLine
    $\x_{\mathrm{pred}} \leftarrow x_\theta(\G_t, t)$\;
    $\mathbf{v}_{\mathrm{pred}} \leftarrow (\x_{\mathrm{pred}} - \G_t) / (1 - t)$\;
    \BlankLine
    $\mathcal{L} \leftarrow || \mathbf{v}_{\mathrm{pred}} - \mathbf{v} ||^2$\;
    $\theta \leftarrow \theta - \eta \nabla_\theta \mathcal{L}$\;
}
\end{algorithm}
\end{minipage}
\hfill
\begin{minipage}[t]{0.48\textwidth}
\begin{algorithm}[H]
\caption{DiGFM Sampling}
\label{alg:sample}

\textbf{Input:} Trained model $x_\theta(\G_t, t)$ and number of sampling steps $K$

Sample number of nodes $n$ from training set\;
Sample $\G_0 \sim \mathcal{N}(0, I_n)$\;
$t \leftarrow 0$\;
\vspace{3pt}
\For{$k=1$ \emph{to} $K$}{
    $\x_{\mathrm{pred}} \leftarrow x_\theta(\G_t, t)$\;
    $\mathbf{v}_{\mathrm{pred}} \leftarrow (\x_{\mathrm{pred}} - \G_t) / (1 - t)$\;
    $\G_{t + 1/K} \leftarrow \G_t + 1/K \cdot \mathbf{v}_{\mathrm{pred}}$\;
    $t \leftarrow t + 1/K$
}
$\G_1 \leftarrow \texttt{quantize}(\G_1)$\;
\textbf{Return:} $\G_1$
\end{algorithm}
\end{minipage}

\textbf{Conditional Generation.} The preceding discussion is for unconditional graph generation. DiGFM also supports conditional generation, where the generated graph is conditioned on a specific numerical property or a categorical label $y$. We adopt the classifier-free guidance (CFG) method for conditioning \cite{ho2021classifier}. During training, we drop the label $y$ for a percentage of the data (e.g., 10\%), replacing it with a null token $\varnothing$ to train an unconditional baseline $\x_\theta(\bm{G}_t, t, \varnothing)$ alongside the conditional model $\x_\theta(\bm{G}_t, t, y)$. During sampling, we apply Algorithm \ref{alg:sample}, except the velocity prediction (line 7) becomes a linear combination of the conditional and unconditional velocities:
\[ \mathbf{v}_{\text{pred}} \leftarrow \mathbf{v}_{\text{uncond}} + s \cdot (\mathbf{v}_{\text{cond}} - \mathbf{v}_{\text{uncond}}), \]
where $s$ is the CFG scale. Setting $s=1$ yields pure conditional generation, while increasing to $s > 1$ provides a stronger conditioning guidance, which improves the generation quality.
\vspace{-6pt}
\section{Experiments}
\vspace{-6pt}

In this section, we conduct experiments to evaluate DiGFM, demonstrating that it outperforms or matches existing models on generating graphs with the target structural statistics while operating significantly faster. We first assess generation performance across general graph benchmarks, molecule datasets, and conditional molecule generation (Section \ref{sec:exp-gen}). Next, we analyze DiGFM's sampling efficiency (Section \ref{sec:exp-sample}). Finally, Section \ref{sec:exp-ablate} presents an ablation study of core components of DiGFM, highlighting the effectiveness of the proposed clean graph prediction approach.

\vspace{-5pt}
\subsection{Experimental Setup}\label{sec:exp-setup}
\vspace{-5pt}

\textbf{Datasets.} We evaluate DiGFM on diverse benchmark datasets for both general graph and molecular generation. For general graph generation, where the focus is on evaluating topology without node or edge attributes, we use four datasets: Community \cite{you2018graphrnn}, Ego \cite{sen2008collective,jo2022score}, SBM \cite{martinkus2022spectre}, and Enzymes \cite{schomburg2004brenda}. These datasets contain two-community graphs, ego networks from Citeseer, stochastic block model graphs, and protein tertiary structure graphs, respectively. For molecular graph generation, which requires modeling both structure and node/edge attributes, we evaluate on the QM9 \cite{ramakrishnan2014quantum} and ZINC250k \cite{irwin2012zinc} datasets. More details about the datasets are in Appendix \ref{app:data}.

\textbf{Evaluation Metrics.} For general graphs, we evaluate structural fidelity by measuring the maximum mean discrepancy (MMD) of four graph statistics between generated graphs and ground-truth graphs in the test set: degree (Deg.), clustering coefficient (Clus.), number of orbits with 4 nodes (Orbit), and the eigenvalues of the graph Laplacian (Spec.) \cite{jo2024graph}. For the SBM dataset, which is the only one with a standard definition of validity, we also report the percentage of valid, unique, and novel graphs (V.U.N.). For the other datasets, we compute the Fréchet Inception Distance (FID) and radial basis function MMD (RBF MMD) to measure the overall generation diversity and fidelity \cite{thompson2022evaluation}.

For molecular graphs, we use four common metrics: the percentage of generated molecules that are valid without post hoc correction (Valid \%), Fréchet ChemNet Distance (FCD) \cite{preuer2018frechet}, neighborhood subgraph pairwise distance kernel (NSPDK) \cite{costa2010fast}, and scaffold similarity (Scaf.) \cite{bemis1996properties}. 

In all tables, we include a ``Training set" reference row. This records the distance between the training and test sets measured by each metric, so it serves as a baseline of the natural statistical distance between the training and test splits drawn from the same graph distribution. An optimal model is expected to achieve metrics close to this ``Training set" baseline.

\textbf{DiGFM Setup.} The hyperparameters of DiGFM for each dataset are given in Appendix \ref{app:details}.

\vspace{-4pt}
\subsection{Graph Generation Performance}\label{sec:exp-gen}
\vspace{-4pt}

\textbf{General Graph Generation.} DiGFM outperforms baselines on the majority of metrics, across all four datasets (Tables \ref{tab:com_ego}-\ref{tab:sbm_enzymes}). DiGFM's low MMD metrics indicate that its generated graphs closely match the structural statistics of the ground-truth graph distribution. Further, the generated graphs are diverse (i.e., the model does not generate the same or similar graphs every time), as shown by the high V.U.N. and low FID and RBF MMD scores. For each metric, results are reported as the mean $\pm$ standard deviation across five sampling runs. We visualize a single graph generated by DiGFM for each dataset in Figure \ref{fig:all-vis}, and additional generations are in Appendix \ref{app:vis}.

Crucially, DiGFM achieves these state-of-the-art generation results using only a fraction of the sampling steps required by baselines. While the majority of baselines require 500 or 1000 steps, DiGFM generates using just 50 steps on the Community, Ego, and Enzymes datasets, and only 40 steps on the SBM dataset. We explore this acceleration further in Section \ref{sec:exp-sample}, demonstrating that DiGFM maintains state-of-the-art performance on Community and Enzymes with as few as 15 steps.

\textbf{Molecular Graph Generation.} DiGFM generates realistic molecules, achieving a higher percentage of chemically valid molecules (Valid \%) and accurate scaffolds (Scaf.) than baselines on QM9 and a higher validity percentage and lower NSPDK on ZINC250k. Unlike the general graph datasets above, the molecular datasets contain node and edge attributes. Specifically, QM9 contains up to 9 heavy atoms of 4 types, and ZINC250k contains up to 38 heavy atoms of 9 types. In both datasets, there are three edge types: single, double, or triple bonds. The results in Table \ref{tab:qm9_zinc} demonstrate that DiGFM not only captures correct structural topologies, but also accurately models complex node and edge attributes.

\vspace{0.2cm}

\begin{figure}[t]
    \centering
    \includegraphics[width=\linewidth]{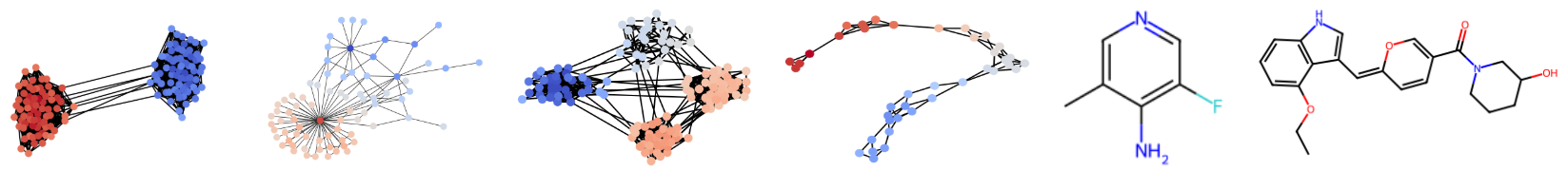}
    \caption{Visualization of a single graph generated by DiGFM on each dataset.}
    \label{fig:all-vis}
    \vspace{-5pt}
\end{figure}

\begin{table}[h!]
\vspace{-3pt}
\centering
\scriptsize
\caption{\small \textbf{Generation results on Community and Ego graphs.} DiGFM outperforms baselines on the majority of metrics, while only using up to 10\% of their sampling steps.}
\vspace{3pt}
\resizebox{0.96\textwidth}{!}{
\begin{tabular}{l ccccc ccccc}\toprule
& \multicolumn{5}{c}{Community (Synthetic)} & \multicolumn{5}{c}{Ego (Real)} \\
\cmidrule(lr){2-6} \cmidrule(lr){7-11}
Model & Deg. $\downarrow$ & Clus. $\downarrow$ & Orbit $\downarrow$ & FID $\downarrow$ & RBF $\downarrow$ & Deg. $\downarrow$ & Clus. $\downarrow$ & Orbit $\downarrow$ & FID $\downarrow$ & RBF $\downarrow$ \\\midrule
Training set & 0.0028 & 0.0128 & 0.0129 & 2.0239 & 0.0134 & 0.0002 & 0.0078 & 0.0060 & 0.4216 & 0.0093 \\
\midrule
GraphRNN \cite{you2018graphrnn} & 0.1440 & 0.0535 & 0.0198 & 8.3869 & 0.1591 & 0.0768 & 1.1456 & 0.1087 & 90.5655 & 0.6827 \\
GRAN \cite{liao2019efficient} & 0.1022 & 0.0894 & 0.0198 & 64.115 & 0.0749 & 0.5778 & 0.3360 & 0.0406 & 489.960 & 0.2633 \\		
GraphCNF \cite{lippe2020categorical} & 0.1129 & 1.2882 & 0.0197 & 29.153 & 0.1341 & 0.1010 & 0.7654 & 0.0820 & 18.7929 & 0.0896 \\
GDSS \cite{jo2022score} & 0.0535 & 0.2072 & 0.0196 & 6.5531 & 0.0443 & 0.8189 & 0.6032 & 0.3315 & 60.6100 & 0.4331 \\
DiscDDPM \cite{haefeli2022diffusion} & 0.1238 & 0.6549 & 0.0246 & 8.6321 & 0.0840 & 0.4613 & 0.1681 & 0.0633 & 42.7994 & 0.1561 \\
DiGress \cite{vignac2023digress} & 0.0409 & \textbf{0.0167} & 0.0298 & 3.4261 & 0.0460 & 0.0708 & \textbf{0.0092} & 0.1205 & 18.6794 & 0.0489 \\
EDGE \cite{chen2023efficient} & 0.0175 & 0.0689 & 0.0198 & 2.2378 & 0.0227 & 0.0579 & 0.1773 & 0.0519 & 15.7614 & 0.0658 \\
\midrule
\multirow{2}{*}{DiGFM (ours)} & \textbf{0.0017} & 0.0220 & \textbf{0.0185} & \textbf{1.3080} & \textbf{0.0186} & \textbf{0.0017} & 0.0641 & \textbf{0.0233} & \textbf{1.9784} & \textbf{0.0212} \\
& \tiny$\pm$0.0022 & \tiny$\pm$0.0019 & \tiny$\pm$0.0012 & \tiny$\pm$0.2304 & \tiny$\pm$0.0009 & \tiny$\pm$0.0002 & \tiny$\pm$0.0088 & \tiny$\pm$0.0032 & \tiny$\pm$0.3972 & \tiny$\pm$0.0012 \\
\bottomrule
\end{tabular}
}
\label{tab:com_ego}
\vspace{-7pt}
\end{table}

\begin{table}[h!]
\centering
\scriptsize
\caption{\small \textbf{Generation results on SBM and Enzymes graphs.} DiGFM outperforms baselines on the majority of metrics, while only using up to 10\% of their sampling steps.}
\vspace{3pt}
\resizebox{0.96\textwidth}{!}{
\begin{tabular}{l ccccc cccc}\toprule
& \multicolumn{5}{c}{SBM (Synthetic)} & \multicolumn{4}{c}{Enzymes (Real)} \\
\cmidrule(lr){2-6} \cmidrule(lr){7-10}
Model & Deg. $\downarrow$ & Clus. $\downarrow$ & Orbit $\downarrow$ & Spec. $\downarrow$ & V.U.N. $\uparrow$ & Deg. $\downarrow$ & Clus. $\downarrow$ & Orbit $\downarrow$ & Avg. $\downarrow$ \\\midrule
Training set & 0.0008 & 0.0332 & 0.0255 & 0.0063 & 0 & 0.003 & 0.020 & 0.017 & 0.014 \\
\midrule
GraphRNN \cite{you2018graphrnn} & 0.0055 & 0.0584 & 0.0785 & 0.0065 & 5 & 0.017 & 0.062 & 0.046 & 0.042 \\
SPECTRE \cite{martinkus2022spectre} & 0.0015 & 0.0521 & 0.0412 & 0.0056 & 53 & - & - & - & - \\
EDP-GNN \cite{niu2020permutation} & \textbf{0.0011} & 0.0552 & 0.0520 & 0.0070 & 35 & 0.023 & 0.268 & 0.082 & 0.124 \\
GDSS \cite{jo2022score} & 0.0212 & 0.0646 & 0.0894 & 0.0128 & 5 & 0.026 & 0.061 & 0.009 & 0.032 \\
EDGE \cite{chen2023efficient} & 0.0279 & 0.1113 & 0.0854 & 0.0251 & 0 & - & - & - & - \\
DiGress \cite{vignac2023digress} & 0.0013 & 0.0498 & 0.0434 & 0.0400 & 74 & 0.030 & 0.060 & 0.011 & 0.034 \\
Graphusion \cite{yang2024graphusion} & - & - & - & - & - & 0.022 & 0.060 & \textbf{0.007} & 0.030 \\
GruM \cite{jo2024graph} & 0.0017 & \textbf{0.0492} & 0.0442 & 0.0071 & 70 & - & - & - & - \\
\midrule
\multirow{2}{*}{DiGFM (ours)} & 0.0026 & 0.0506 & \textbf{0.0379} & \textbf{0.0050} & \textbf{80.5} & \textbf{0.004} & \textbf{0.035} & 0.010 & \textbf{0.017} \\
 & \tiny$\pm$0.0012 & \tiny$\pm$0.0009 & \tiny$\pm$0.0065 & \tiny$\pm$0.0008 & \tiny$\pm$3.7 & \tiny$\pm$0.0004 & \tiny$\pm$0.0007 & \tiny$\pm$0.0032 & \tiny$\pm$0.0014 \\
\bottomrule
\end{tabular}
\vspace{-7pt}
}
\label{tab:sbm_enzymes}
\end{table}

\begin{table}[h!]
\vspace{-7pt}
\centering
\scriptsize
\caption{\small \textbf{Generation results on molecule datasets.} DiGFM achieves competitive performance on both QM9 and ZINC250k datasets, while only using up to 15.6\% of their sampling steps.}
\vspace{3pt}
\resizebox{0.96\textwidth}{!}{
\begin{tabular}{l cccc cccc}\toprule
& \multicolumn{4}{c}{QM9 (Real)} & \multicolumn{4}{c}{ZINC250k (Real)} \\
\cmidrule(lr){2-5} \cmidrule(lr){6-9}
Model & Valid \% $\uparrow$ & FCD $\downarrow$ & NSPDK $\downarrow$ & Scaf. $\uparrow$ & Valid \% $\uparrow$ & FCD $\downarrow$ & NSPDK $\downarrow$ & Scaf. $\uparrow$ \\\midrule
Training set & 100 & 0.040 & 0.0001 & 0.9719 & 100 & 0.062 & 0.0001 & 0.8395 \\
\midrule
MoFlow \cite{zang2020moflow} & 91.36 & 4.467 & 0.0169 & 0.1447 & 63.11 & 20.931 & 0.0455 & 0.0133 \\
GraphAF \cite{shi2020graphaf} & 74.43 & 5.625 & 0.0207 & 0.3046 & 68.47 & 16.023 & 0.0442 & 0.0672 \\
GraphDF \cite{luo2021graphdf} & 93.88 & 10.93 & 0.0636 & 0.0978 & 90.61 & 33.546 & 0.1770 & 0.0000 \\
EDP-GNN \cite{niu2020permutation} & 47.52 & 2.680 & 0.0046 & 0.3270 & 82.97 & 16.737 & 0.0485 & 0.0000 \\
GDSS \cite{jo2022score} & 95.72 & 2.900 & 0.0033 & 0.6983 & 97.01 & 14.656 & 0.0195 & 0.0467 \\
DiGress \cite{vignac2023digress} & 98.19 & \textbf{0.095} & 0.0003 & 0.9353 & 94.99 & 3.482 & 0.0021 & 0.4163 \\
GruM \cite{jo2024graph} & 99.69 & 0.108 & \textbf{0.0002} & 0.9449 & 98.65 & 2.257 & 0.0015 & \textbf{0.5299} \\
DeFoG \cite{qin2025defog} & 98.90 & 0.260 & 0.0004 & 0.9192 & 96.65 & \textbf{2.123} & 0.0022 &  0.4245 \\
\midrule
\multirow{2}{*}{DiGFM (ours)} & \textbf{99.82} & 0.106 & 0.0005 & \textbf{0.9450} & \textbf{98.73} & 2.206 & \textbf{0.0014} & 0.5195 \\
& \tiny$\pm$ 0.05 & \tiny$\pm$ 0.005 & \tiny$\pm$ 0.00003 & \tiny$\pm$ 0.0035 & \tiny$\pm$ 0.19 & \tiny$\pm$ 0.035 & \tiny$\pm$ 0.00005 & \tiny$\pm$ 0.0216 \\
\bottomrule
\end{tabular}
}
\label{tab:qm9_zinc}
\end{table}

\vspace{5pt}

\textbf{Conditional Graph Generation.} With its direct graph prediction design, DiGFM natively supports conditional generation. We evaluate this capability in Table \ref{tab:qm9-cond} by generating QM9 molecules conditioned on the dipole moment $\mu$, the Highest Occupied Molecular Orbital (HOMO), and both. In all three cases, DiGFM achieves the best overall results, with the highest validity and lowest mean absolute error (MAE) from the target $\mu$ and/or HOMO values.

Additionally, the table highlights the effectiveness of the classifier-free guidance mechanism for DiGFM. The top half of the table shows the model performance without guidance, and the bottom half shows the performance using the optimal guidance scale. (We ablate the optimal guidance scale in Section \ref{sec:exp-ablate}.) When guidance is used, DiGFM performs much better on the MAE metric, which shows that the model is successfully steered toward generating molecules with the desired dipole moment or HOMO. At the same time, DiGFM maintains a high validity under guidance. 

\begin{table}[h!]
\vspace{-5pt}
\centering
\scriptsize
\caption{\textbf{Conditional generation on QM9 molecules.} DiGFM outperforms baselines on generation of QM9 molecules conditioned on the dipole moment $\mu$, the Highest Occupied Molecular Orbital (HOMO), and both.}
\vspace{3pt}
\resizebox{0.96\textwidth}{!}{
\begin{tabular}{l c cc cc cc}
\toprule
& & \multicolumn{2}{c}{Dipole Moment ($\mu$)} & \multicolumn{2}{c}{HOMO} & \multicolumn{2}{c}{Both ($\mu$ \& HOMO)} \\
\cmidrule(lr){3-4} \cmidrule(lr){5-6} \cmidrule(lr){7-8}
Model & With guidance? & MAE ($\downarrow$) & Valid \% ($\uparrow$) & MAE ($\downarrow$) & Valid \% ($\uparrow$) & MAE ($\downarrow$) & Valid \% ($\uparrow$) \\
\midrule
DiGress \cite{vignac2023digress} & No & 1.68 & 91.5 & 0.95 & 91.5 &  1.32 & 91.5 \\
FreeGress \cite{ninniri2024classifier} & No & 1.03 & 86.7 & 0.45 & 92.0 & 0.83 & 82.3 \\
DiGFM (Ours) & No & 1.67 & 96.4 & 0.26 & 96.1 & 0.60 & \textbf{96.7} \\
\midrule
DiGress \cite{vignac2023digress} & Yes & 0.80 & 82.5 & 0.61 & 91.2 &  1.11 & 86.7 \\
FreeGress \cite{ninniri2024classifier} & Yes & 0.74 & 83.7 & 0.32 & 90.1 & 0.68 & 83.4 \\
DiGFM (Ours) & Yes & \textbf{0.57} & \textbf{96.5} & \textbf{0.19} & \textbf{96.7} & \textbf{0.44} & 95.7 \\
\bottomrule 
\end{tabular}
}
\label{tab:qm9-cond}
\vspace{-2pt}
\end{table}

\subsection{Improved Sampling Efficiency}\label{sec:exp-sample}
\begin{figure}[h!]
    \vspace{-4pt}
    \centering
    \includegraphics[width=\linewidth]{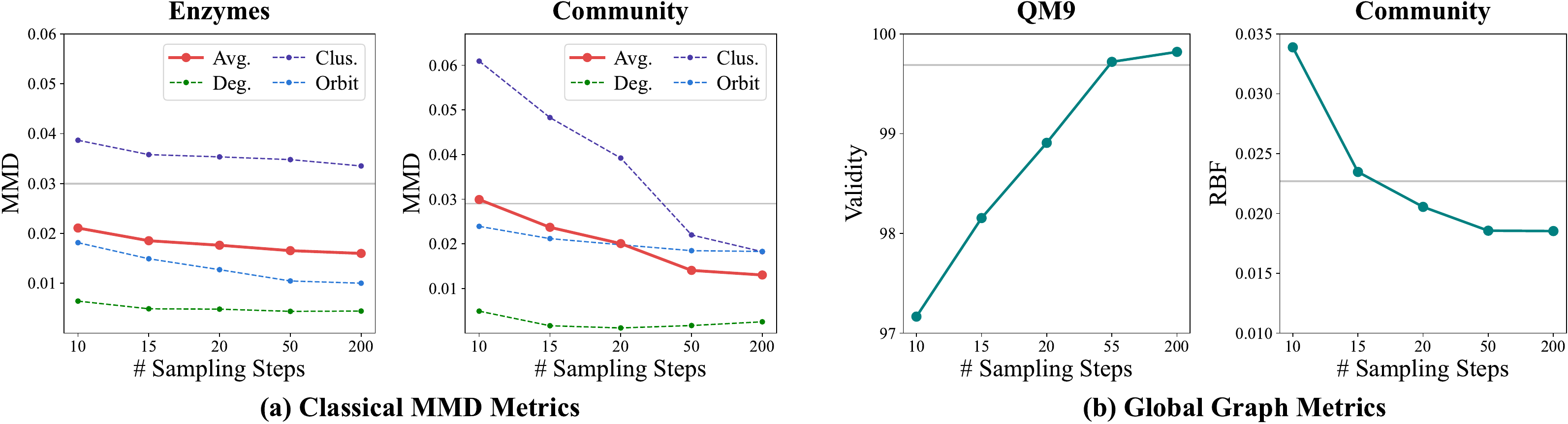}
    \caption{\textbf{Sampling efficiency of DiGFM. (a)} DiGFM's average MMD (red line) outperforms the state-of-the-art (gray line) in only 15 steps. \textbf{(b)} The global graph metrics of Validity and RBF MMD surpass the state-of-the-art at just 55 and 20 steps, respectively.}
    \label{fig:15steps}
    \vspace{-5pt}
\end{figure}

\vspace{-5pt}
We demonstrate that DiGFM is highly efficient at inference time: it outperforms previous models on both classical MMD and global graph metrics while using only 2.5\% to 15.6\% of their sampling steps. As Figure \ref{fig:15steps}(a) shows, DiGFM rapidly converges to the ground-truth graph topology, requiring only 15 sampling steps to outperform baselines. For example, a 15-step DiGFM achieves an average MMD of 0.019 on the Enzymes dataset (surpassing Graphusion's 600-step MMD of 0.030) and 0.024 on the Community dataset (surpassing DiGress's 500-step MMD of 0.029). Beyond structural accuracy, Figure \ref{fig:15steps}(b) confirms that DiGFM maintains overall generation diversity and graph validity even with this small number of steps. Notably, this step efficiency translates directly to wall-clock speedups: as Table \ref{tab:inference-time} shows, DiGFM achieves a $5.3\times$ to $257\times$ speedup over all six baselines. Further, DiGFM remains $5.3\times$ to $12\times$ faster even when compared to a reduced-step version of DeFoG (a state-of-the-art discrete model that was the most efficient in previous literature).

\begin{table}[h!]
\tiny
\centering
\caption{\textbf{Inference time comparison.} Time (in seconds) required to generate a single graph on an A100-SXM4-80GB GPU. DiGFM is more efficient than all six baseline models.}
\vspace{3pt}
\label{tab:inference-time}
\setlength{\tabcolsep}{3.1pt}
\begin{tabular}{@{}l*{13}{c}@{}}
\toprule
 & \multicolumn{7}{c}{\textbf{Inference Time (sec/graph)}} & \multicolumn{6}{c}{\textbf{Speedup}} \\
\cmidrule(lr){2-8} \cmidrule(lr){9-14}
Dataset & GruM & DiGress & GDSS & DisCo & CatFlow & DeFoG & DiGFM (ours) & vs.\ GruM & vs.\ DiGress & vs.\ GDSS & vs.\ DisCo & vs.\ CatFlow & vs.\ DeFoG \\
\midrule
SBM      & 12.75 & 19.92 & 12.39 & 0.995 & 2.11 & 0.714 & \textbf{0.136} & ${94\times}$  & ${146\times}$ & ${91\times}$ & ${7.3\times}$ & ${16\times}$ & ${5.3\times}$ \\
QM9      & 12.06 & 9.09  & 11.60 & 1.88 & 1.59 & 0.578 & \textbf{0.047} & ${257\times}$ & ${193\times}$ & ${247\times}$ & ${40\times}$ & ${34\times}$ & ${12\times}$ \\
ZINC250k & 12.59 & 11.34 & 12.36 & 2.37 & 2.47 & 0.835 & \textbf{0.073} & ${172\times}$  & ${155\times}$ & ${169\times}$ & ${32\times}$ & ${34\times}$ & ${11\times}$ \\
\bottomrule
\end{tabular}
\end{table}

\vspace{-2pt}
\subsection{Ablations}\label{sec:exp-ablate}
\vspace{-5pt}

Finally, we ablate several core components of DiGFM.

\textbf{Effectiveness of Direct Graph Prediction.} We demonstrate that direct graph prediction ($\x$-prediction) is critical for accurate topology generation. Table \ref{tab:exp-pred} compares DiGFM against two variants with identical backbones trained to directly output either velocity or noise. The models were trained on the Ego dataset. The velocity variant optimizes the flow matching loss: $\mathcal{L} = \mathbb{E}_{t, \x, \e} \; \Vert{} \mathbf{v}_\theta(\z_t, t) - (\x - \e) \Vert{}^2$. The noise variant outputs predicted noise $\e_\theta(\z_t, t)$, which is converted to velocity via $\mathbf{v}_\theta(\z_t, t) = (\z_t - \e_\theta(\z_t, t)) / t$ prior to computing the same loss. 

Crucially, although all the models in Table \ref{tab:exp-pred} optimize mathematically equivalent flow matching objectives, parameterizing the output in clean data space ($\x$-space) yields significantly better performance. This aligns with the manifold hypothesis \cite{chapelle2006semi}: Gaussian noise is high-dimensional and off-manifold, making it difficult for the model to learn. Velocity ($\mathbf{v} = \x - \e$) is also difficult because it combines the clean graph with unstructured noise. In contrast, the clean graph $\x$ itself lies on a structured, low-dimensional data manifold, providing the most stable and easily learnable target for the model.

\definecolor{bad}{RGB}{255,235,235}
\definecolor{good}{RGB}{235,255,235}

\begin{table}[h!]
\vspace{-7pt}
\scriptsize
\centering
\caption{Predicting the clean graph $\x$ outperforms predicting the velocity $\mathbf{v}$ or noise $\e$.}
\vspace{3pt}
\begin{NiceTabular}{ccccccc}
\toprule
Prediction Target & Deg. $\downarrow$ & Clus. $\downarrow$ & Orbit $\downarrow$ & Spec. $\downarrow$ & FID $\downarrow$ & RBF $\downarrow$ \\
\midrule
\cellcolor{good} Clean graph $\x$ & \cellcolor{good} 0.0017 & \cellcolor{good} 0.0641 & \cellcolor{good} 0.0233 & \cellcolor{good} 0.0050 & \cellcolor{good} 1.9784 & \cellcolor{good} 0.0212 \\
\cellcolor{bad} Velocity $\mathbf{v}$ & \cellcolor{bad} 0.0022& \cellcolor{bad} 0.0870& \cellcolor{bad} 0.0268& \cellcolor{bad} 0.0064& \cellcolor{bad} 4.1700& \cellcolor{bad} 0.0335 \\
\cellcolor{bad} Noise $\e$ & \cellcolor{bad} 0.0028 & \cellcolor{bad} 0.0882 & \cellcolor{bad} 0.0262 & \cellcolor{bad} 0.0101 & \cellcolor{bad} 6.7882 & \cellcolor{bad} 0.0407 \\
\bottomrule
\end{NiceTabular}
\label{tab:exp-pred}
\end{table}

\vspace{4pt}

\textbf{Ablation on DiGFM Architectural Designs.} Table \ref{tab:ablate-k} evaluates two architectural designs introduced in Section \ref{sec:method-arch}. First, we ablate the number of top Laplacian eigenvectors ($k$) used as the target node attributes on the Enzymes dataset. Performance generally improves as $k$ increases, with $k=4$ achieving the best overall results. Second, we evaluate the inclusion of higher-order structural features. Using both the first- and second-order powers of the adjacency matrix ($\bm{E}$ and $\bm{E}^2$) outperforms using $\bm{E}$ alone and yields results comparable to adding $\bm{E}^3$. Thus, we adopt $\bm{E}$ and $\bm{E}^2$ as an optimal setting to achieve strong performance with only a minimal increase in computational complexity, as measured by the number of seconds to generate one graph on an A100-SXM4-80GB GPU.

\begin{table}[h!]
\vspace{-7pt}
\centering
\scriptsize
\caption{
Ablation on DiGFM architectural designs.
}
\vspace{3pt}
\begin{tabular}{l c c c c c c c c} %
\toprule
Model & Deg. $\downarrow$ & Clus. $\downarrow$ & Orbit $\downarrow$ & Spec. $\downarrow$ & FID $\downarrow$ & RBF $\downarrow$ & Timing (sec/graph) $\downarrow$ \\
\specialrule{\lightrulewidth}{\aboverulesep}{0pt}
\rowcolor{gray!20} \multicolumn{8}{l}{\emph{Ablation on \# Laplacian eigenvectors as node features}} \\
DiGFM (1 eigenvector) & 0.0072 & 0.0393 & 0.0239 & 0.0175 & 9.5679 & 0.0349 & \textbf{0.0296} \\
DiGFM (2 eigenvectors) & 0.0076 & 0.0369 & 0.0119 & 0.0150 & 9.7655 & 0.0333 & 0.0298 \\
DiGFM (3 eigenvectors) & 0.0063 & 0.0360 & 0.0125 & 0.0143 & 7.2446 & 0.0305 & 0.0299 \\
DiGFM (4 eigenvectors) & \textbf{0.0044} & \textbf{0.0348} & \textbf{0.0105} & \textbf{0.0120} & \textbf{6.6211} & \textbf{0.0287} & 0.0302 \\
\specialrule{\lightrulewidth}{\aboverulesep}{0pt}
\rowcolor{gray!20} \multicolumn{8}{l}{\emph{Ablation on higher-order structural features}} \\
DiGFM (with $\bm{E}$) & 0.0057 & 0.0355 & 0.0114 & 0.0134 & 6.6419 & \textbf{0.0277} & \textbf{0.0298} \\
DiGFM (with $\bm{E}, \bm{E}^2$) & \textbf{0.0044} & 0.0348 & 0.0105 & \textbf{0.0120} & \textbf{6.6211} & 0.0287 & 0.0302 \\
DiGFM (with $\bm{E}, \bm{E}^2, \bm{E}^3$) & 0.0051 & \textbf{0.0347} & \textbf{0.0104} & \textbf{0.0120} & 6.7209 & 0.0294 & 0.0304 \\
\bottomrule 
\end{tabular}
\label{tab:ablate-k}
\end{table}

\vspace{4pt}

\begin{wrapfigure}{r}{0.4\textwidth}
    \vspace{-\intextsep}
    \centering
    \includegraphics[width=\linewidth]{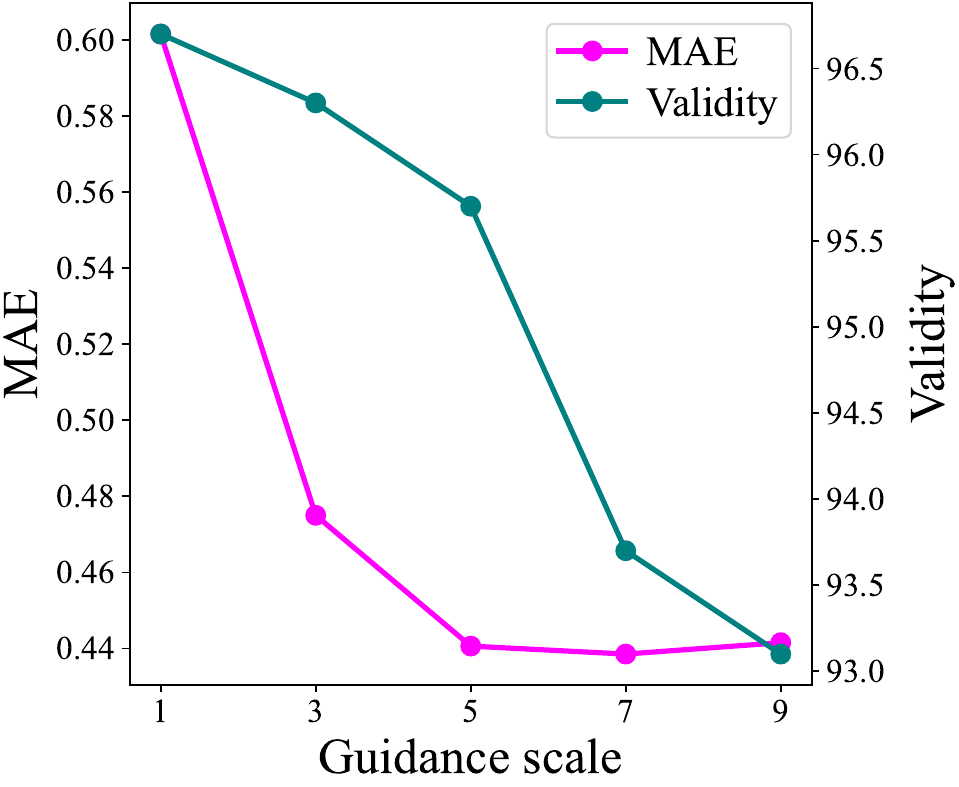}
    \caption{Ablation on the guidance scale}
    \label{fig:cfg-ablate}
\end{wrapfigure}

\textbf{Ablation on Guidance Scale.} Figure \ref{fig:cfg-ablate} demonstrates the trade-off between property adherence and molecular validity when adjusting DiGFM's guidance scale for conditional generation on QM9. Property adherence is measured by the MAE, where a lower value indicates the generated molecule's $\mu$ and HOMO match the target values more closely. Molecular validity reflects the percentage of chemically valid outputs. As the guidance scale increases, the MAE improves, but validity decreases. As shown in Figure \ref{fig:cfg-ablate}, a guidance scale of 3 or 5 provides an optimal balance between low MAE and high validity. Since this scale can be flexibly adjusted at inference time without retraining, the optimal scale for each scenario can be identified by sweeping the candidate values.

\WFclear
\vspace{-4pt}
\section{Conclusion}
\vspace{-4pt}

In this work, we present Direct Graph Flow Matching (DiGFM), a novel generative model designed to overcome the limitations of standard noise-prediction methods in diffusion and flow models. By leveraging a continuous flow-matching paradigm integrated with direct graph prediction, DiGFM accurately learns the clean graph’s topological properties. A key aspect of our approach is the separate formulation of the model output space (clean graph) and the loss space (velocity), with a transformation between them. Extensive experiments demonstrate that DiGFM outperforms or matches state-of-the-art baselines across general graph benchmarks and molecular datasets, while achieving a 5.3$\times$ to 257$\times$ speedup in inference time.

\bibliographystyle{unsrtnat}
\bibliography{reference}

\newpage

\etocdepthtag.toc{appendix}
\etocsettagdepth{main}{none} 
\etocsettagdepth{appendix}{subsection} 
\etocsettocstyle{\section*{Appendix Contents}}{}
\tableofcontents
\vspace{1em}

\appendix
\addtocontents{toc}{\protect\setlength{\cftbeforesecskip}{2pt}}

\vspace{-4pt}
\section{Model Implementation Details}\label{app:details}

The hyperparameters of DiGFM are given in Table \ref{tab:hyperparameters}. Our codebase is at \url{https://github.com/Susie-Lu/DiGFM}.

\begin{table}[h!]
\vspace{-7pt}
\small
\centering %
\caption{Configurations of DiGFM on each dataset.}
\vspace{3pt}
\begin{tabular}{l c c c} %
\toprule
Config & Community, Ego, Enzymes, SBM & QM9 & ZINC250k \\
\midrule
\# params & 7.1M & 8.2M & 10.8M \\
Depth & 8 & 8 & 12 \\
MLP dims & $X$: 128, $E$: 64, $y$: 128 & $X$: 256, $E$: 256, $y$: 128 & $X$: 256, $E$: 128, $y$: 128 \\
Hidden dims & $d_x$: 256, $d_e$: 64, $d_y$: 64 & $d_x$: 256, $d_e$: 128, $d_y$: 64 & $d_x$: 256, $d_e$: 64, $d_y$: 64 \\
Num heads & 8 & 8 & 8 \\
FFN dims & $X$: 256, $E$: 64, $y$: 256 & $X$: 256, $E$: 256, $y$: 128 & $X$: 256, $E$: 128, $y$: 128 \\
Init features & Top Laplacian eigenvectors & One-hot encodings & One-hot encodings \\
\midrule
Batch size & 8 & 2048 & 1024 \\
Optimizer & AdamW & AdamW & AdamW \\
Learning rate & $2.0 \times 10^{-4}$ & $2.0 \times 10^{-4}$ & $2.0 \times 10^{-4}$ \\
Weight decay & $10^{-12}$ & $10^{-12}$ & $10^{-12}$ \\
EMA half-life & 30K graphs & 400K graphs & 300K graphs \\
\bottomrule 
\end{tabular}
\label{tab:hyperparameters}
\end{table}

We provide additional details about DiGFM and its training that do not appear in the main text.

\textbf{Stabilizing Training via EMA.} To stabilize training, we employ an adaptive Exponential Moving Average (EMA) \cite{karras2022elucidating} on the model weights. Specifically, at step $s$ in the training loop, the EMA decay rate is $\beta = 0.5^{\text{bsz} /\text{halflife}}$, where $\text{bsz}$ is the batch size. Here, $\text{halflife} = \min(\text{tgt\_halflife}, 0.05 \cdot \text{n\_graphs})$, where $\text{tgt\_halflife}$ is the specified target halflife and $\text{n\_graphs}=\text{bsz}\cdot s$ is the number of graphs that the model has processed by step $s$ in the training loop. With this approach, the EMA model weights update rapidly early in training and change more slowly later in training.

\textbf{Output Bounding.} Since DiGFM directly predicts the clean graph rather than unbounded, off-manifold Gaussian noise, we can explicitly constrain the model's output to match the valid domain of the target data. We apply these bounding operations after the final transformer block:
\begin{itemize}
\item \textbf{Edge Attributes ($\bm{E}$):} We apply a sigmoid activation, $\bm{E}_{\text{out}} = \text{sigmoid}(\bm{E}_{\text{pred}})$, to bound predicted edge values to the interval $[0, 1]$. 
\item \textbf{Node Attributes ($\bm{X}$):} For categorical node attributes, we apply a node-wise softmax activation, $\bm{X}_{\text{out}} = \text{softmax}(\bm{X}_{\text{pred}})$, so that the output is a valid probability distribution. For continuous attributes where the ground-truth node attributes come from Laplacian eigenvectors, we normalize each predicted node attribute vector to an L2 norm of $1$, which ensures that it matches the norm of the ground-truth eigenvectors.
\end{itemize}

\textbf{Laplacian Spectral Node Features.} For graphs without given node attributes, we set the target node attribute matrix $\bm{X}$ to consist of the eigenvectors of the combinatorial Laplacian $L = D - A$, where $A$ is the adjacency matrix and $D$ is the diagonal degree matrix. Specifically, we extract and $L_2$-normalize the $k$ eigenvectors corresponding to the smallest non-zero eigenvalues. Additionally, to account for the sign ambiguity in standard eigensolvers, we apply a random sign-flipping augmentation during training. At each training step and for each graph, we independently sample a sign vector $\sigma \in \{-1, 1\}^k$ and multiply it with the precomputed eigenvectors to yield the input $\bm{X}$. This ensures the model learns representations that are invariant to arbitrary eigenvector signs.

\textbf{Additional Discussion on $\x$-prediction outperforming $\mathbf{v}$-prediction.} Although $\mathbf{v}$ and $\x$ are mathematically related by the simple linear transformation $\mathbf{v}(\z_t, t) = \tfrac{1}{1 - t} (\x(\z_t, t) - \z_t)$ in equation \eqref{eq:x-to-v}, $\z_t$ in this equation is a high-dimensional noisy quantity, which is difficult for the model to represent internally. Predicting $\mathbf{v}$ is significantly harder than predicting $\x$ because it requires the model to internally process $\z_t$:
\begin{itemize}
    \item With $\x$-prediction, the high-dimensional quantity $\z_t$ in the transformation equation is supplied \emph{outside} the network. The network only needs to learn to represent the low-dimensional clean graph $\x$.
    \item With $\mathbf{v}$-prediction, the network must learn the dependence on the high-dimensional, noisy quantity $\z_t$ \emph{internally}.
\end{itemize}
This difference becomes important when the dimension of the noisy graph $\z_t$ approaches or exceeds the model’s representational capacity. If the network contains any structural bottlenecks (such as narrower linear layers), some of the high-dimensional noise information required for $\z_t$ would be discarded, which degrades the performance of $\mathbf{v}$-prediction. In contrast, with 
$\x$-prediction, the target output is a lower-dimensional clean graph, so a finite-capacity network can learn it more easily.

\vspace{5pt}

\textbf{Detailed Versions of Training and Sampling Algorithms.} We provide a more detailed version of Algorithms \ref{alg:train} and \ref{alg:sample} in the main text, where we explicitly write out the node matrix $\bm{X}_t$ and edge matrix $\bm{E}_t$ rather than using the short-hand notation of $\G_t$. Note that for undirected graphs, we draw one Gaussian noise value for each unordered node pair $i<j$ and set both $(i,j)$ and $(j,i)$ entries of $\bm{E}_0$ to this value; this ensures that the noisy edge matrix $\bm{E}_0$ is symmetric with a zero diagonal.
\vspace{-6pt}

{\begin{minipage}[t]{0.48\textwidth}
\small
\begin{algorithm}[H]
\caption{DiGFM Training}
\label{alg:train-extended}

\textbf{Input:} Model $x_\theta((\bm{X}_t, \bm{E}_t), t)$

\For{\emph{\textbf{each step}}}{
    Sample $\G_1=(\bm{X}_1,\bm{E}_1)$ from dataset \;
    $n \leftarrow$ number of nodes of $\G_1$ \;
    Sample $t \sim \mathcal{U}[0, 1]$\;
    Sample $\bm{X}_0 \sim \mathcal{N}(0,1)^{n \times d_x}$\;
    Sample $\bm{E}_0 \sim \mathcal{N}(0,1)^{n \times n \times d_e}$\;    
    \BlankLine
    $\bm{X}_t \leftarrow t \cdot \bm{X}_1 + (1-t) \cdot \bm{X}_0$\;
    $\bm{E}_t \leftarrow t \cdot \bm{E}_1 + (1-t) \cdot \bm{E}_0$\;
    $\mathbf{v}_{\bm{X}} \leftarrow (\bm{X}_1 - \bm{X}_t) / (1 - t)$\;
    $\mathbf{v}_{\bm{E}} \leftarrow (\bm{E}_1 - \bm{E}_t) / (1 - t)$\;
    \BlankLine
    $\bm{X}_{\mathrm{pred}}, \bm{E}_{\mathrm{pred}} \leftarrow x_\theta((\bm{X}_t,\bm{E}_t), t)$\;
    $\mathbf{v}_{\bm{X},\mathrm{pred}} \leftarrow (\bm{X}_{\mathrm{pred}} - \bm{X}_t) / (1 - t)$\;
    $\mathbf{v}_{\bm{E},\mathrm{pred}} \leftarrow (\bm{E}_{\mathrm{pred}} - \bm{E}_t) / (1 - t)$\;
    \BlankLine
    $\mathcal{L} \leftarrow \| \mathbf{v}_{\bm{X},\mathrm{pred}} - \mathbf{v}_{\bm{X}} \|^2 + \| \mathbf{v}_{\bm{E},\mathrm{pred}} - \mathbf{v}_{\bm{E}} \|^2$\;
    $\theta \leftarrow \theta - \eta \nabla_\theta \mathcal{L}$\;
}
\end{algorithm}
\end{minipage}
\hfill
\small
\begin{minipage}[t]{0.48\textwidth}
\begin{algorithm}[H]
\caption{DiGFM Sampling}
\label{alg:sample-extended}

\textbf{Input:} Trained model $x_\theta((\bm{X}_t, \bm{E}_t), t)$ and number of sampling steps $K$

Sample number of nodes $n$ from dataset\;
Sample $\bm{X}_0 \sim \mathcal{N}(0,1)^{n \times d_x}$\;
Sample $\bm{E}_0 \sim \mathcal{N}(0,1)^{n \times n \times d_e}$\; 
$t \leftarrow 0$\;
\vspace{6pt}
\For{$k=1$ \emph{to} $K$}{
    $\bm{X}_{\mathrm{pred}}, \bm{E}_{\mathrm{pred}} \leftarrow x_\theta((\bm{X}_t,\bm{E}_t), t)$\;
    \BlankLine
    $\mathbf{v}_{\bm{X},\mathrm{pred}} \leftarrow (\bm{X}_{\mathrm{pred}} - \bm{X}_t) / (1 - t)$\;
    $\mathbf{v}_{\bm{E},\mathrm{pred}} \leftarrow (\bm{E}_{\mathrm{pred}} - \bm{E}_t) / (1 - t)$\;
    \BlankLine
    $\bm{X}_{t + 1/K} \leftarrow \bm{X}_t + 1/K \cdot \mathbf{v}_{\bm{X},\mathrm{pred}}$\;
    $\bm{E}_{t + 1/K} \leftarrow \bm{E}_t + 1/K \cdot \mathbf{v}_{\bm{E},\mathrm{pred}}$\;
    \BlankLine
    $t \leftarrow t + 1/K$
}
\vspace{6pt}
$\G_1 \leftarrow \texttt{quantize}(\bm{X}_1, \bm{E}_1)$\;
\textbf{Return:} $\G_1$
\end{algorithm}
\end{minipage}}

\textbf{Permutation Equivariance.} DiGFM is permutation equivariant with respect to node orderings: permuting the nodes of an input graph results in the same permutation of the model's predicted node and edge matrices. This property follows from the permutation-equivariant graph transformer backbone \cite{vignac2023digress,dwivedi2021generalization} and is preserved by our flow-matching formulation.

\vspace{-5pt}
\section{Dataset and Evaluation Details}\label{app:data}

\begin{wraptable}[5]{r}{0.4\textwidth}
\vspace{-4.3\intextsep}
\scriptsize
\centering
\caption{Dataset statistics.}
\vspace{-3pt}
\begin{tabular}{l c c}
\toprule
Dataset & Min nodes & Max nodes \\
\midrule
Community & 60 & 160 \\
Ego & 50 & 399 \\
SBM & 44 & 187 \\
Enzymes & 10 & 125 \\
QM9 & 1 & 9 \\
ZINC250k & 6 & 38 \\
\bottomrule
\end{tabular}
\label{tab:node-counts}
\end{wraptable}

\textbf{Dataset Details.} We provide more details about the six datasets used. The Community dataset consists of 500 two-community graphs \cite{you2018graphrnn}. The Ego dataset consists of 757 3-hop ego networks extracted from the Citeseer network \cite{sen2008collective,jo2022score}. The SBM dataset consists of 200 graphs from the stochastic block model \cite{martinkus2022spectre}. The Enzymes dataset consists of 587 protein graphs representing the protein tertiary structures of enzymes from the BRENDA database \cite{schomburg2004brenda}. The QM9 dataset consists of 133,885 molecules with up to 9 heavy atoms of 4 types \cite{ramakrishnan2014quantum}. The ZINC250k dataset consists of 249,455 molecules with up to 38 heavy atoms of 9 types \cite{irwin2012zinc}. In both QM9 and ZINC250k, there are three edge types: single, double, or triple bonds.

\textbf{Evaluation Details for General Graph Datasets.} To evaluate the MMD, FID, and RBF MMD metrics for Community and Ego datasets, we follow the protocol from \cite{chen2023efficient,thompson2022evaluation}. We generate 128 graphs from the model and compare against the test set, which contains 102 graphs for Community and 152 graphs for Ego. For FID and RBF MMD, the feature extractor used is a random Graph Isomorphism Network (GIN) with 3 rounds of graph propagation and a node embedding size of 35 recommended by \cite{thompson2022evaluation}, and we average the metrics over 5 random seeds. To evaluate the metrics for the Enzymes and SBM datasets, we follow \cite{jo2022score} and \cite{martinkus2022spectre}, respectively. We generate 117 and 40 graphs, respectively, from the model and compare against the graphs in the test set. The number of generated graphs equals the number of graphs in the test set. Our evaluation code for MMD and V.U.N. follows \cite{jo2024graph}, and our evaluation code for FID and RBF MMD follows \cite{thompson2022evaluation}.

\textbf{Evaluation Details for Molecular Datasets.} We evaluate model performance using the four standard metrics used in \cite{jo2024graph}: Valid \% is the percentage of generated molecules that are valid without post hoc correction such as valency correction or edge resampling. Fréchet ChemNet Distance (FCD) \cite{preuer2018frechet} is the Fréchet distance between the feature vectors of generated molecules and the test set. Neighborhood subgraph pairwise distance kernel (NSPDK) \cite{costa2010fast} measures the Maximum Mean Discrepancy (MMD) between the graph distributions of the generated and test sets by comparing their local neighborhood subgraphs. Scaffold similarity (Scaf.) \cite{bemis1996properties} is the cosine similarity of the frequencies of the Bemis-Murcko scaffolds, which evaluates the model's ability to generate similar substructures as those in the test set.

To evaluate these metrics, we generate 10,000 molecules and compare them against the test set. We use the preprocessed dataset and train-test split from \cite{jo2022score,jo2024graph}: specifically, we kekulized the molecules using the RDKit library \cite{rdkit} and removed hydrogen atoms following the standard procedure. 

\textbf{Evaluation Details for Conditional Molecular Generation.} We evaluate model performance using the standard metrics in \cite{ninniri2024classifier}. The dipole moment is measured in units of Debye, and the HOMO is measured in units of eV. They are computed using RDKit \cite{rdkit} and psi4 \cite{turney2012psi4}. To create the target conditioning vectors $\bm{y}$ for dipole moment and/or HOMO, we randomly sample 100 molecules from the dataset and compute their target properties. Then, we use each vector 10 times to perform conditional generation, so a total of 1000 molecules are generated. We evaluate performance via the Mean Absolute Error (MAE) metric, which is
$$ \frac{1}{1000} \sum_{i=1}^{100} \sum_{j=1}^{10} | \bm{y}_i - \hat{\bm{y}}_{i,j} |, $$
where $\bm{y}_i$ is the target property of the $i$th molecule in the dataset and $\hat{\bm{y}}_{i,j}$ is the target property of the $j$th generated molecule that was generated using $\bm{y}_i$ as conditioning.

\textbf{Details about Figure \ref{fig:15steps}.} For each dataset in Figure \ref{fig:15steps}, we train our DiGFM model and vary the number of sampling steps at inference time. During training, the timestep $t$ is capped at a maximum of $0.99$ (our default setting). When DiGFM is sampled using 10, 15, 20, or 50 steps in the figure, the ODE solver never queries the velocity field for $t > 0.99$, so the training cap on $t$ does not affect generation. However, when using 200 sampling steps, the ODE solver takes its last step at $t = 199/200 = 0.995$, which slightly exceeds the range of training times. Despite this single out-of-distribution step, DiGFM remains robust and maintains strong performance.

In Figure \ref{fig:15steps}, the horizontal gray lines represent the corresponding metrics of the state-of-the-art baseline model. For example, on QM9, 55-step DiGFM achieves a higher validity than GruM's 1000-step value. On the Community dataset, 20-step DiGFM achieves a lower RBF MMD than EDGE's 128-step value.

\textbf{Details about Inference Time Measurements in Table \ref{tab:inference-time}.} All inference times were measured on the same NVIDIA A100-SXM4-80GB GPU. We compared DiGFM against six baseline models: GruM \cite{jo2024graph}, DiGress \cite{vignac2023digress}, GDSS \cite{jo2022score}, DisCo \cite{xu2024discrete}, CatFlow \cite{eijkelboom2024vfm}, and DeFoG \cite{qin2025defog}. We used the model configurations and code from the official repositories for all baselines except CatFlow. Since no official code repository is available for CatFlow, we implemented it using the pseudocode and hyperparameter settings in the paper. The reported times are averaged across 30 generated graphs, excluding two initial warm-up graphs.

\textbf{Details about Baselines in Table \ref{tab:exp-pred}.} The noise-prediction baseline in Table \ref{tab:exp-pred} outputs the predicted noise $\e_\theta(\z_t, t)$, which is converted to velocity via $\mathbf{v}_\theta(\z_t, t) = (\z_t - \e_\theta(\z_t, t)) / t$ prior to computing the flow matching loss. To prevent division by zero, we clip the denominator $t$ to a minimum value of 0.01. This is similar to the numerical stabilization used in our DiGFM model, which computes velocity via $\mathbf{v}_\theta(\z_t, t) = (\x_\theta(\z_t, t) - \z_t) / (1 - t)$. In this case, we bound the denominator $1 - t$ from below at 0.01 (equivalent to clipping $t$ to a maximum of 0.99).

\section{Additional Ablations}\label{app:ablate}

\begin{figure}[h!]
    \centering
    \includegraphics[width=0.7\linewidth]{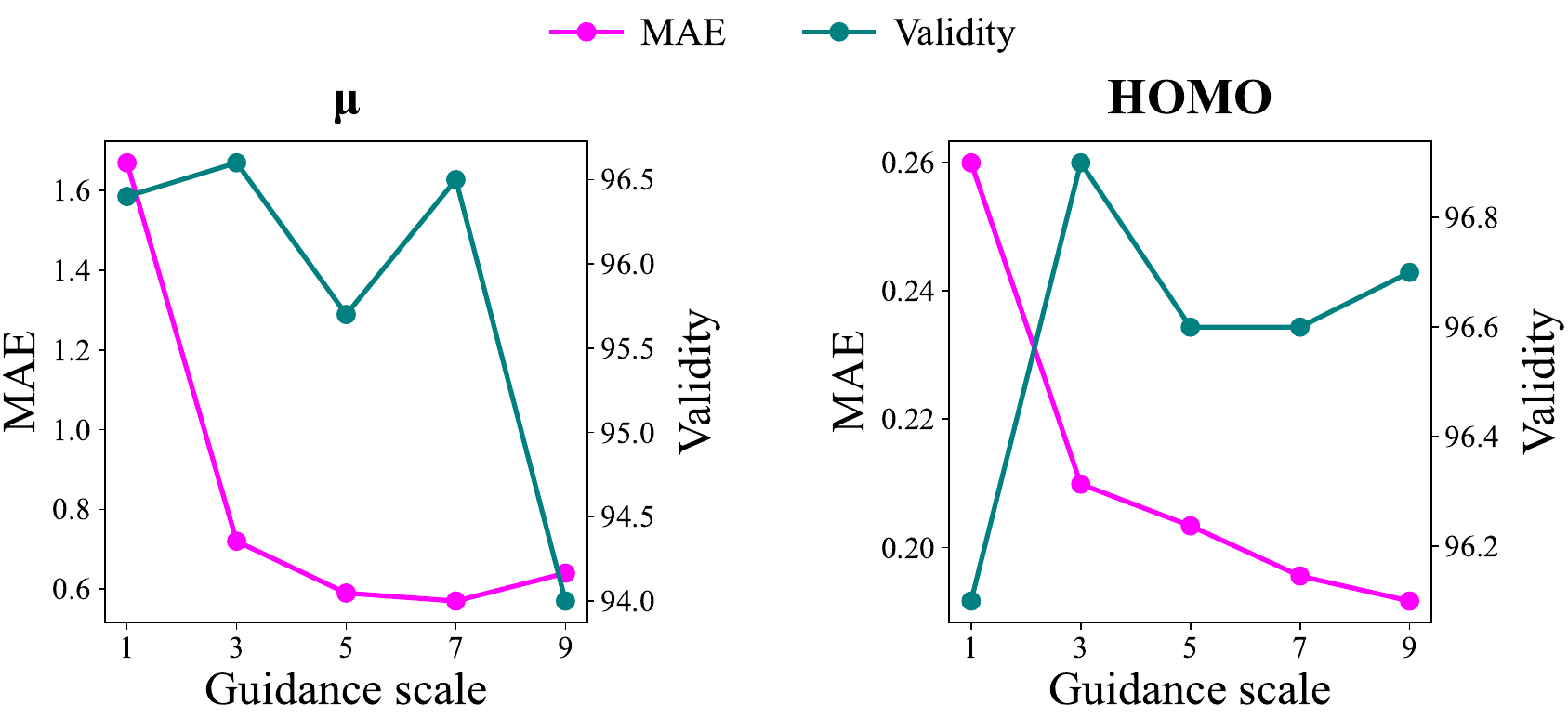}
    \caption{\textbf{Ablation on guidance scale.} Optimal performance is achieved at a scale of 7 for the $\mu$-conditioned model (left), and at scales of 3 and 9 for the HOMO-conditioned model (right).}
    \label{fig:cfg-ablate-more}
\end{figure}

\textbf{Ablation on Guidance Scale.} Figure \ref{fig:cfg-ablate} in the main text ablates the guidance scale for the DiGFM model conditioned jointly on the dipole moment ($\mu$) and HOMO. Figure \ref{fig:cfg-ablate-more} below provides the ablation results for DiGFM conditioned on $\mu$ only (left) and on HOMO only (right). For the $\mu$-conditioned model, a guidance scale of 7 achieves the lowest MAE and highest validity. For the HOMO-conditioned model, the guidance scales of 3 and 9 both yield strong performance.

\textbf{Ablation on the Range of Flow Matching Timesteps.} To prevent division by zero in the $(1-t)$ denominator of equation \eqref{eq:x-to-v}, we cap the flow matching time step $t$ to be at most $(K_{\text{max}}-1)/K_{\text{max}}$, where $K_{\text{max}}$ is the maximum intended number of sampling steps at inference. This does not affect the sampling process, because at inference, a $K$-step Euler ODE solver only evaluates the velocity field at time steps $t \in \{0, 1/K, \dots, (K-1)/K\}$, so it never queries $t > (K_{\text{max}}-1)/K_{\text{max}}$. 

Table \ref{tab:ablate-clip} provides empirical validation. We train DiGFM on the Enzymes dataset with the time step capped at 0.95, 0.99, and 0.995, keeping all other settings identical. We sample from the model using 50 steps. As shown, the model's performance remains consistent across all three cap values. 

\begin{table}[h!]
\vspace{-5pt}
\centering
\scriptsize
\caption{
Ablation for the cap on the maximum flow matching time step used when training DiGFM. DiGFM's performance remains consistent across all three cap values.
}
\vspace{3pt}
\begin{tabular}{ccccccc} %
\toprule
Cap on Time Step & Deg. $\downarrow$ & Clus. $\downarrow$ & Orbit $\downarrow$ & Spec. $\downarrow$ & FID $\downarrow$ & RBF $\downarrow$ \\
\midrule
0.95 & 0.0047 & \textbf{0.0334} & 0.0133 & 0.0129 & \textbf{5.8646} & \textbf{0.0276} \\
0.99 & \textbf{0.0044} & 0.0348 & \textbf{0.0105} & \textbf{0.0120} & 6.6211 & 0.0287 \\
0.995 & 0.0053 & 0.0341 & 0.0113 & 0.0128 & 6.9350 & 0.0307 \\
\bottomrule 
\end{tabular}
\label{tab:ablate-clip}
\end{table}

\section{Additional Sampling Efficiency Comparisons}

\textbf{Additional Inference Time Comparison.} In the main text, Table \ref{tab:inference-time} shows that DiGFM is faster than all six baseline models at inference time. The DeFoG model \cite{qin2025defog} in Table \ref{tab:inference-time} is its accelerated, reduced-step version. We also report the inference time for the full version of DeFoG in Table \ref{tab:inference-time-extra} below.

In the timing comparison, DiGFM uses 40, 50, and 50 sampling steps on the SBM, QM9, and ZINC250k datasets, respectively. The reduced-step version of DeFoG uses 50 steps across all three datasets, and its full version uses 1000, 500, and 1000 steps, respectively.

Although reduced-step DeFoG and our proposed DiGFM use a similar number of sampling steps, DiGFM achieves faster wall-clock inference times. This is because DiGFM (as a continuous flow matching model) only requires a fast Euler ODE solver update, while DeFoG (as a discrete flow matching model) must build a discrete transition process and sample categorical states at every step, which leads to slower inference.

\begin{table}[h!]
\vspace{-5pt}
\scriptsize
\centering
\caption{\textbf{Inference time comparison.} Time (in seconds) required to generate a single graph on an A100-SXM4-80GB GPU. DiGFM is more efficient than both the reduced-step and full versions of DeFoG.}
\vspace{3pt}
\label{tab:inference-time-extra}

\setlength{\tabcolsep}{3.1pt}
\begin{tabular}{@{}l*{6}{c}@{}}
\toprule
 & \multicolumn{3}{c}{\textbf{Inference Time (sec/graph)}} & \multicolumn{2}{c}{\textbf{Speedup}} \\
\cmidrule(lr){2-4} \cmidrule(lr){5-6}
Dataset & Full DeFoG & Reduced-step DeFoG & DiGFM (ours) & vs.\ Full DeFoG & vs.\ Reduced-step DeFoG \\
\midrule
SBM      & 10.86 & 0.714 & \textbf{0.136} & $80\times$ & ${5.3\times}$ \\
QM9      & 5.71 & 0.578 & \textbf{0.047} & $121\times$ & ${12\times}$ \\
ZINC250k & 13.34 & 0.835 & \textbf{0.073} & $183\times$ & ${11\times}$ \\
\bottomrule
\end{tabular}
\vspace{8pt}
\end{table}

\textbf{Scalability Comparison.} Like other representative graph diffusion and flow models (e.g., the discrete models such as DiGress \cite{vignac2023digress}, DisCo \cite{xu2024discrete}, Cometh \cite{siraudin2024cometh}, and DeFoG \cite{qin2025defog}, as well as the continuous model GruM \cite{jo2024graph}), DiGFM has an $\mathcal{O}(N^2)$ complexity, where $N$ is the number of nodes in the generated graph. The reason is that these models all rely on dense adjacency matrices to represent all $N(N-1)$ node pairs, and they use a graph transformer backbone \cite{vignac2023digress,dwivedi2021generalization} with $\mathcal{O}(N^2)$ time and memory complexity per layer. In this paper, our focus is on improving sampling efficiency within this standard regime, and extending DiGFM to achieve subquadratic complexity is an orthogonal direction for future work.

\begin{figure}[h!]
    \centering
    \includegraphics[width=0.7\linewidth]{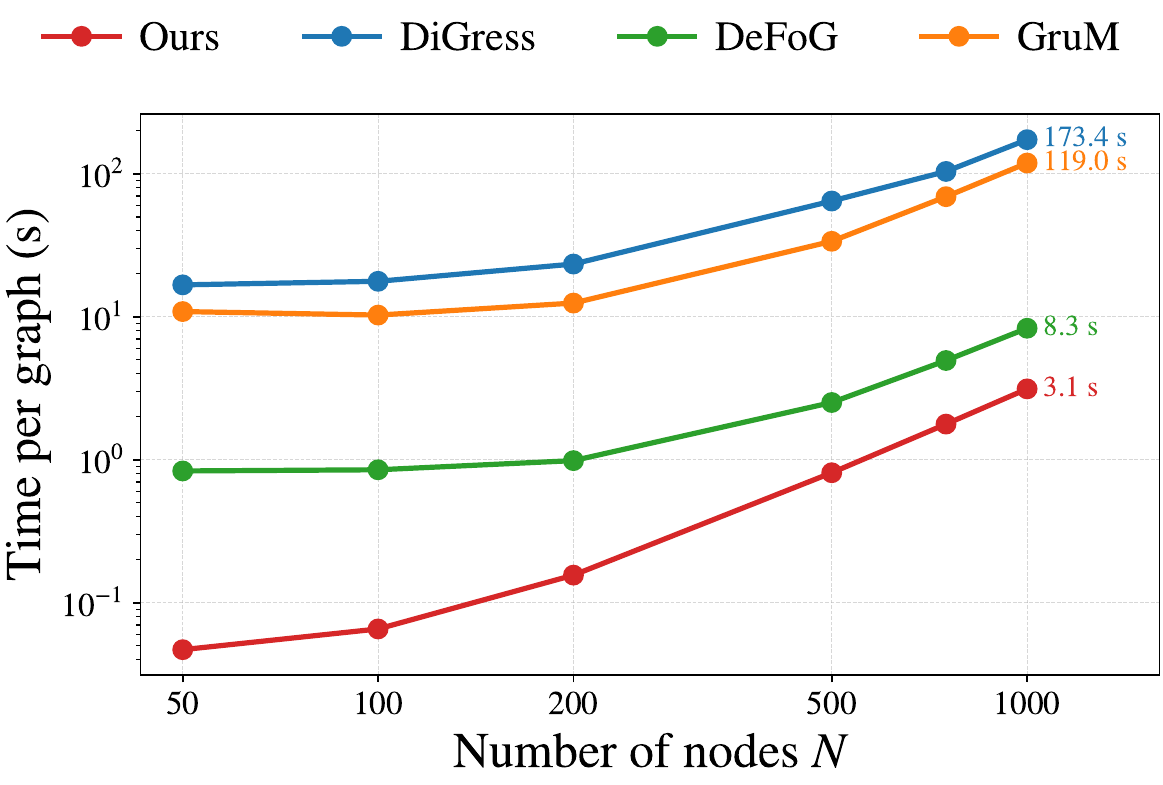}
    \caption{\textbf{Inference time scaling comparison.} DiGFM maintains a significant speed advantage over baseline models (DeFoG, DiGress, and GruM) as the number of generated nodes $N$ scales up to 1000.}
    \label{fig:scale}
    \vspace{15pt}
\end{figure}

Figure \ref{fig:scale} demonstrates that DiGFM maintains high efficiency and continues to have a significant inference speed advantage over baselines as graph sizes scale, despite the quadratic complexity. To evaluate scalability up to 1000 nodes, we compared DiGFM against DeFoG, DiGress, and GruM (using SBM hyperparameters) across sizes $N \in \{50, 100, 200, 500, 1000\}$. Notably, at 1000 nodes, DiGFM requires just 3.13 seconds per graph, which is 2.7$\times$, 56$\times$, and 38$\times$ faster than DeFoG, DiGress, and GruM, respectively. While the generated 1000-node graphs are not meaningful in this setup (as the SBM dataset caps at 187 nodes), the timing comparison is valid for comparing scalability. To measure graph generation within the training range, we note that generating a 399-node Ego graph on an A100-SXM4-80GB GPU takes DiGFM only 0.610 seconds.

\vspace{0.4cm}

\newpage

\section{Additional Related Work}

\textbf{Comparison with Few-step Graph Diffusion/Flow Models.} Standard graph diffusion and flow matching models often need 500 to 1000 sampling steps to generate a clean graph \cite{vignac2023digress,jo2022score}, making inference slow. Recent work has focused on reducing this step count to speed up generation \cite{xu2024discrete,siraudin2024cometh,qin2025defog}. For example, DeFoG \cite{qin2025defog} is a state-of-the-art discrete flow matching model that generates high-quality graphs in just 50 steps. Our proposed DiGFM has several main differences from DeFoG.

First, DiGFM is the first \emph{continuous} flow matching model that is able to generate high-quality graphs in the few-step regime (15--50 sampling steps). Our approach of using direct graph prediction to make few-step generation possible is completely different from DeFoG's discrete designs, such as nonlinear sample time distortion and changes to the transition rate matrix.

Second, even though both models use a similar number of steps, DiGFM is 5.3$\times$ to 12$\times$ faster in wall-clock inference time (Table \ref{tab:inference-time}). This is because DiGFM's continuous formulation only requires a fast Euler solver update, while DeFoG has to build a discrete transition process and sample categorical states at every step.

\section{Generated Graphs}\label{app:vis}
In the following Figures \ref{fig:enzymes}-\ref{fig:zinc}, we visualize samples generated from DiGFM for the Enzymes, SBM, Community, Ego, QM9, and ZINC250k datasets.

\vspace{0.8cm}

\begin{figure}[h!]
\centering
\includegraphics[width=0.95\textwidth]{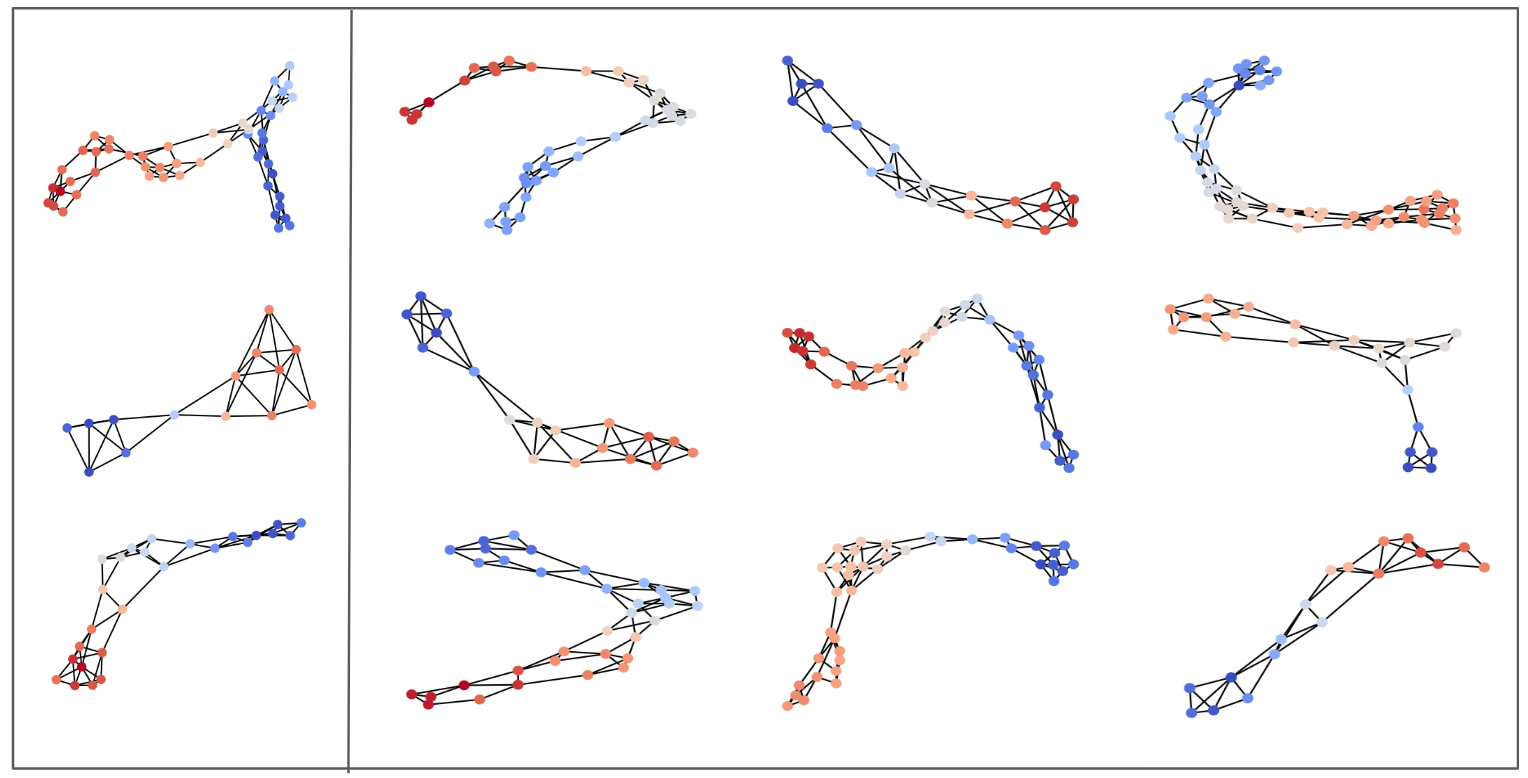}
\caption{Graphs from the Enzymes test set (left) vs. samples generated by DiGFM (right).}
\label{fig:enzymes}
\end{figure}

\vspace{1.2cm}

\begin{figure}[h!]
\centering
\includegraphics[width=0.95\textwidth]{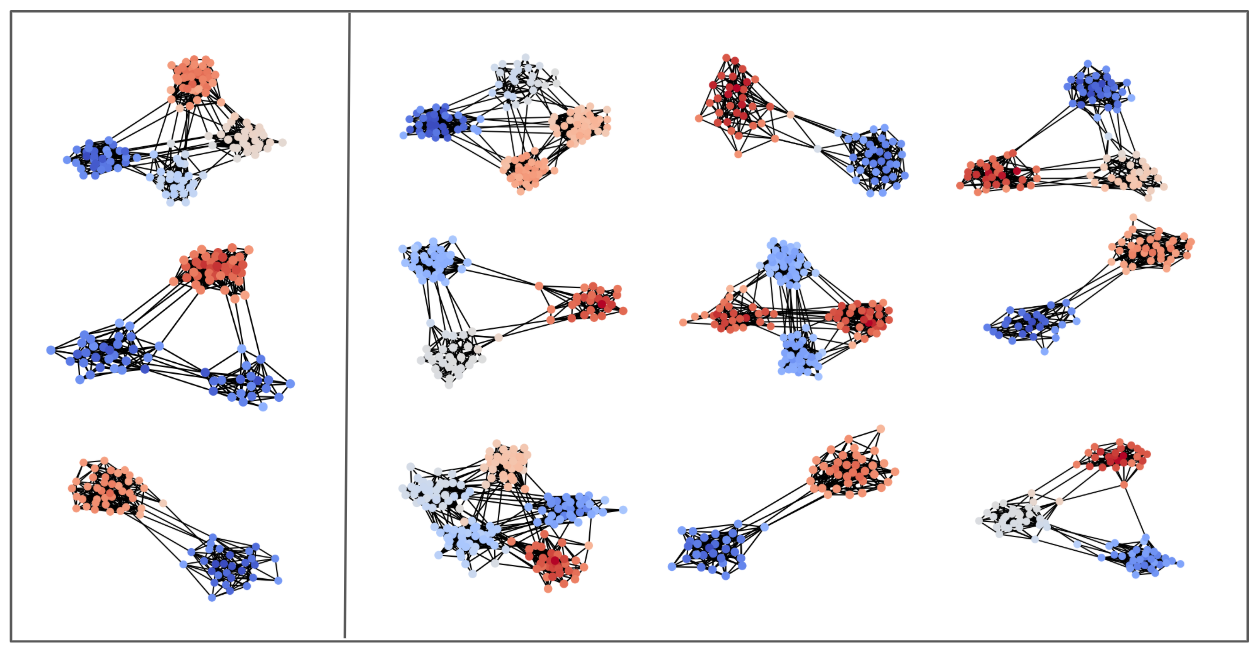}
\caption{Graphs from the SBM test set (left) vs. samples generated by DiGFM (right).}
\label{fig:sbm}
\end{figure}

\begin{figure}[h!]
\centering
\includegraphics[width=0.95\textwidth]{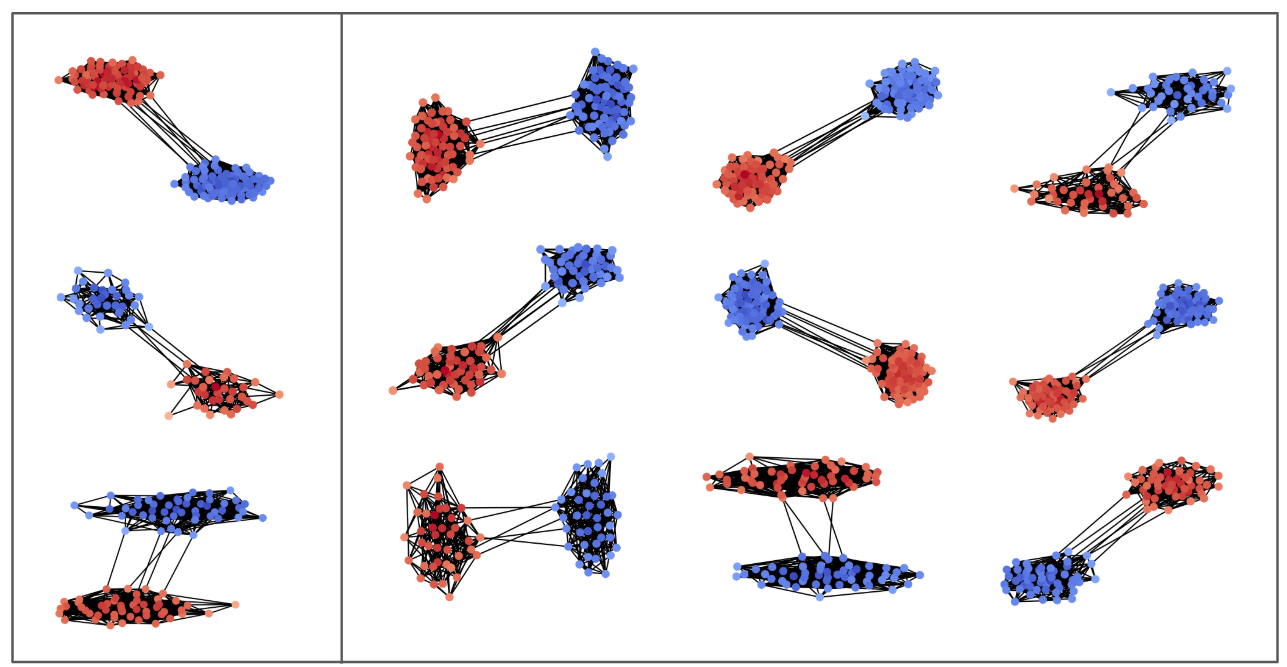}
\caption{Graphs from the Community test set (left) vs. samples generated by DiGFM (right).}
\label{fig:community}
\end{figure}

\begin{figure}[h!]
\centering
\includegraphics[width=0.95\textwidth]{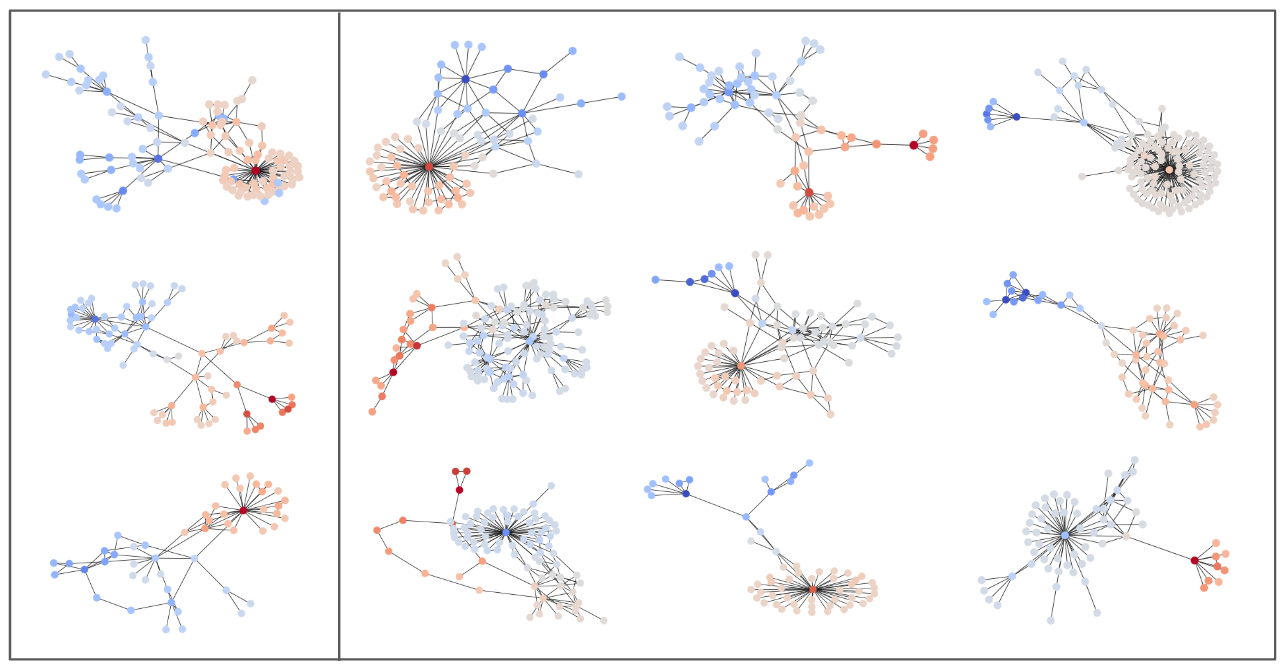}
\caption{Graphs from the Ego test set (left) vs. samples generated by DiGFM (right).}
\label{fig:ego}
\end{figure}

\begin{figure}[h!]
\centering
\includegraphics[width=0.95\textwidth]{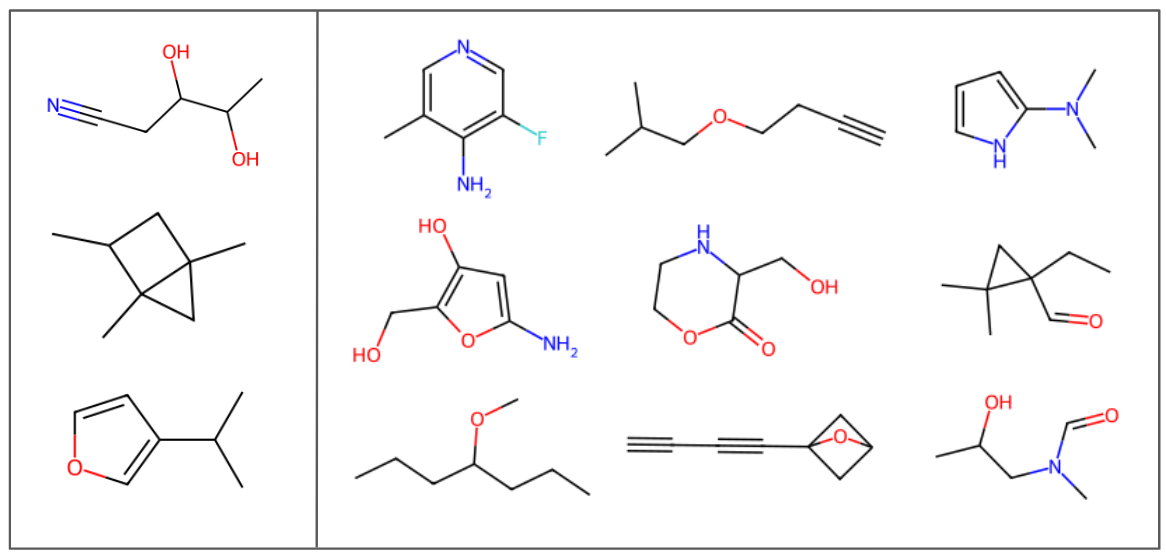}
\caption{Graphs from the QM9 test set (left) vs. samples generated by DiGFM (right).}
\label{fig:qm9}
\end{figure}

\begin{figure}[h!]
\centering
\includegraphics[width=0.96\textwidth]{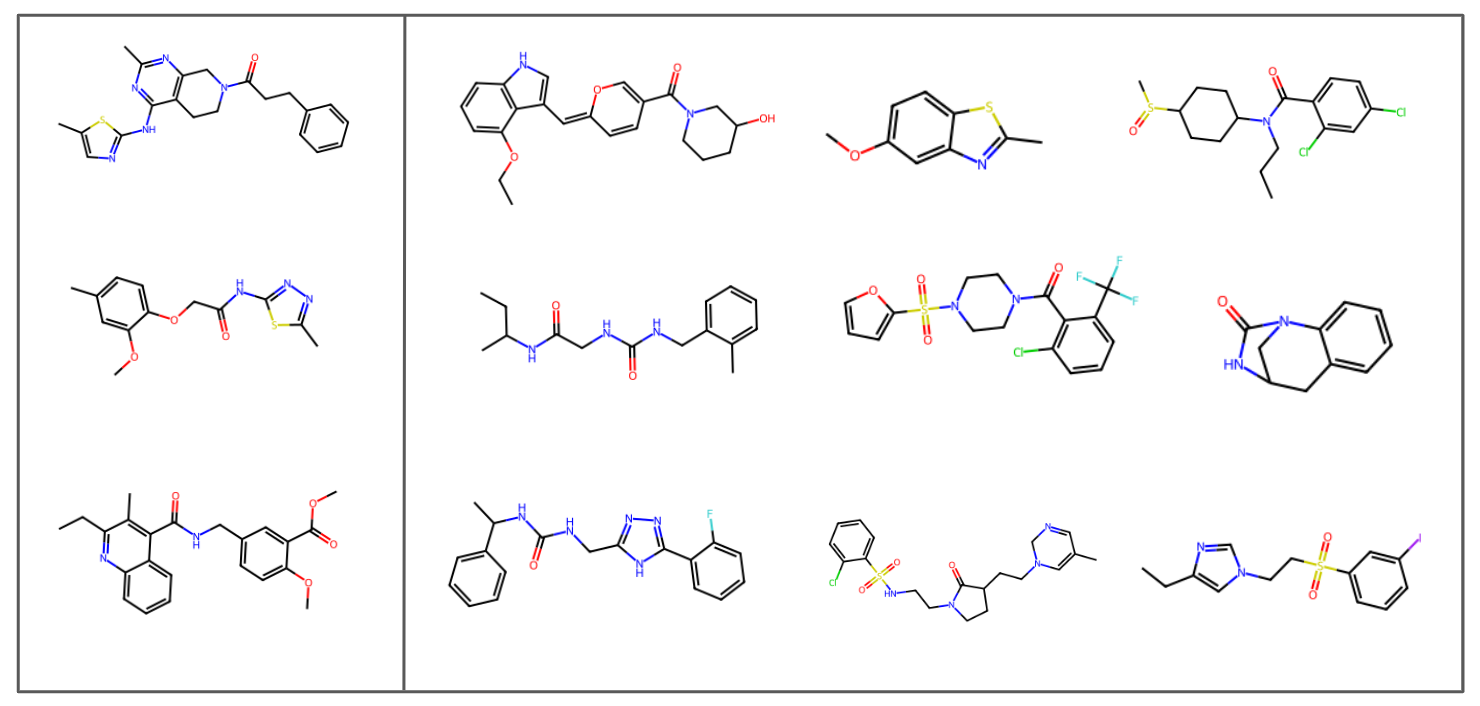}
\caption{Graphs from the ZINC250k test set (left) vs. samples generated by DiGFM (right).}
\label{fig:zinc}
\end{figure}

\end{document}